\documentclass[letterpaper, 10 pt, conference]{ieeeconf}
\IEEEoverridecommandlockouts    % to override locked commands
\usepackage[noadjust]{cite}

\usepackage{multirow}
\usepackage{lipsum}
\usepackage{amsmath}
\usepackage{amssymb}
\usepackage[ruled,vlined]{algorithm2e}
\usepackage{graphicx}
\usepackage{xcolor}
\newtheorem{problem}{Problem}
\usepackage{tcolorbox}
\usepackage{float}
\usepackage{listings}
\usepackage{hyperref}
\usepackage[ font=footnotesize]{caption}
\usepackage{amsfonts}
\usepackage{dsfont}

\usepackage{xcolor}
\usepackage{arydshln}
\usepackage{bbm}
\usepackage{adjustbox}
\usepackage{caption}
\usepackage{subcaption}
\usepackage[skip=2pt]{caption}
\graphicspath{{./Figures/}}
\title{\LARGE \bf Language-Driven Cost Optimization for Autonomous Driving}

\author{
 	\parbox{\textwidth}{%
		\centering
 		Diego Martinez-Baselga$^{1}$, Khaled Mustafa$^{1}$, Javier Alonso-Mora$^{1}$%
 	}%
 	\thanks{This research has been supported by the Dutch Research Council NWO-NWA, within the “Acting under uncertainty” (ACT) project (Grant No. NWA.1292.19.298).}%
 	\thanks{$^{1}$ Cognitive Robotics Department, TU Delft, The Netherlands {\tt\small \{d.martinezbaselga, k.a.mustafa, j.alonsomora\}@tudelft.nl}}%
}

\begin{document}
	
	\maketitle
	\thispagestyle{empty}
	\pagestyle{empty}
	
	%%%%%%%%%%%%%%%%%%%%%%%%%%%%%%%%%%%%%%%%%%%%%%%%%%%%%%%%%%%%%%%%%%
	\begin{abstract}
The driving behavior of autonomous vehicles is typically governed by the cost function of their motion planner, which encodes objectives such as speed tracking, smoothness, lane keeping, and collision avoidance. However, tuning the parameters that shape this cost function is a challenging task that requires technical expertise, limiting the vehicle’s ability to adapt to evolving traffic scenarios or end-user preferences. This work presents a language-driven framework for adaptive cost design in autonomous driving. A Large Language Model (LLM) interprets structured scenario descriptions and natural language user queries to generate the parameters applied to a risk-aware Model Predictive Path Integral (MPPI) controller. The system incorporates a \textit{human-in-the-loop} validation stage in which the proposed behavioral changes are described in non-technical language and confirmed prior to deployment. Users may additionally provide feedback either before or after deployment, enabling iterative refinement of the vehicle’s motion behavior. The framework is evaluated across multiple queries in realistic driving scenarios  to assess its effectiveness. Simulation results demonstrate that our method successfully induces behavioral changes that align with the intended requirements in an intuitive manner, thereby bridging the gap between intelligent vehicle control systems and end users.
	\end{abstract}
	
	\section{Introduction}
Motion planning plays a central role in autonomous driving systems. It is responsible for computing safe and dynamically feasible trajectories that account for the vehicle’s dynamics, surrounding agents, %environmental constraints, 
and context. 
In many systems, motion behavior is encoded through a weighted cost function that is minimized by an optimal control or sampling-based planner, such as Model Predictive Control (MPC) or Model Predictive Path Integral (MPPI) control~\cite{Gratzer2025AgileML,Zhang2025IntegratingDW}. The relative weighting of individual cost terms determines the resulting driving style. However, these weights are typically tuned offline through empirical evaluation and remain fixed during deployment. As a result, the planner lacks the ability to adapt its behavior online to scenario-specific contextual information, such as traffic density, road geometry, or interaction patterns with other agents.

%In addition to environmental context, motion 
Passenger driving preferences may vary %across passengers and situations. D
depending on urgency, comfort requirements, physical condition, or traffic conditions. While such preferences can be naturally expressed in everyday language, translating qualitative instructions into quantitative modifications of a motion planner’s cost function remains challenging. This difficulty is particularly pronounced for non-expert users who are unfamiliar with control-theoretic concepts or cost function design. Recent advances in Large Language Models (LLMs)~\cite{comanici2025gemini,hurst2024gpt} enable structured reasoning over contextual and linguistic inputs, making them promising interfaces between human intent and parameterized optimization planning frameworks.

In this work, we propose a language-driven motion planning framework that adapts cost function parameters based on scenario descriptions and natural language user queries. An LLM module interprets contextual and behavioral inputs to adjust the parameters of a cost function %that governs the motion planner. The resulting cost function is 
optimized by an MPPI framework, which computes real-time control inputs for the vehicle. To enhance reliability and user-trust, we incorporate a \textit{human-in-the-loop} validation stage. Before deployment, the proposed behavioral modifications are translated into non-technical language and presented to the user for confirmation. The user can provide feedback to refine or correct the suggested parameter adjustments, thereby reducing the risk of misinterpretation and improving alignment between intent and execution. Furthermore, we maintain conversational context across successive interactions, enabling users to introduce additional behavioral refinements during execution. This design allows intuitive and flexible control of a risk-aware autonomous driving planner without requiring expertise in optimal control or cost design. 

The contributions of this work are threefold. First, we introduce an \textbf{adaptive language-driven motion planning framework} where the vehicle behavior adapts to diverse environments and user requirements via a rich risk-aware cost function optimized through MPPI, enabling an LLM module to adjust parameters on the fly while ensuring real-time safety. Second, we propose a \textbf{human-in-the-loop validation and refinement} scheme that allows users to review, confirm, and iteratively refine LLM-generated parameter updates through a structured feedback loop to mitigate misunderstandings. Finally, we demonstrate the effectiveness of our framework through \textbf{evaluation in a realistic simulator} across various driving scenarios.

    \section{Related work}
\begin{figure*}
    \centering
    \includegraphics[trim={0cm 16cm 0cm 0cm},clip, width=0.8\linewidth]{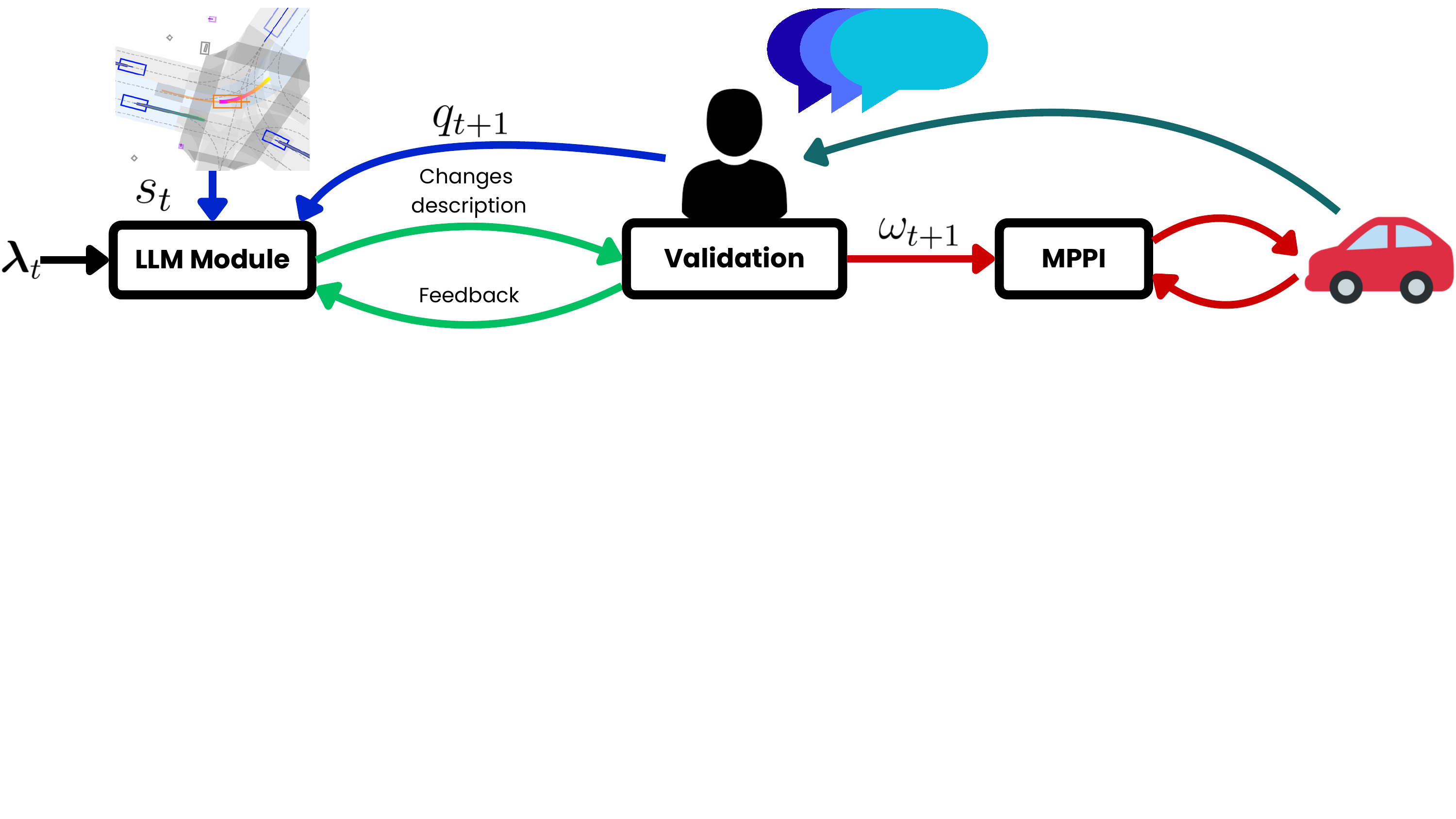}
    \caption{Overview of the system. The LLM is queried by a new scenario or the user (blue arrows). It proposes modifications that must be validated by the user (green arrows). The new weights are applied in the MPPI (red arrows). The user may continue sending queries as the vehicles navigates (gray arrow).}
    \label{fig:overview}
\end{figure*}
Recent advancements have explored the use of Large Language Models (LLMs) and Vision Language Models (VLMs) into autonomous driving pipelines to enhance vehicles perception, planning and decision-making capabilities \cite{cui2024survey,tian2026large}. Several approaches employ LLMs and VLMs for high-level decision making in autonomous driving, reasoning about traffic scenes and outputting discrete driving decisions, such as accelerating, braking, or changing lanes \cite{sima2024drivelm,fu2024drive,wen2024dilu,jin2024surrealdriver,shinn2023reflexion,jiang2024koma,lin2025sandra}. %By leveraging large-scale pretraining, these systems enhance situational reasoning and common-sense understanding. 
Vision-Language Action (VLA) models, in contrast, have the ability to directly output continuous control commands %, mapping perception inputs to low-level actions through large-scale data-driven training 
from multimodal inputs \cite{brohan2023rt,zitkovich2023rt,kim2024openvla,zhao2025cot}. While such approaches demonstrate the potential of foundation models for behavioral reasoning and action generation, they often lack explicit guarantees on control stability, constraint satisfaction, and safety structure. Moreover, achieving reliable real-time feasibility in complex urban scenarios remains challenging. 

A parallel line of work explores the use of LLMs in continuous control settings. 
Prior works use language models to propose waypoints for autonomous driving~\cite{shao2024lmdrive}, generate reward functions~\cite{han2024words,huang2024vlm} or translate high-level human instructions into control objectives in robotics settings~\cite{liang2023code,brohan2023can}. These works highlight the potential of language models as interfaces between human intent and control systems, yet they often operate in simplified environments or lack structured safety formulations tailored to urban driving.

More closely related are methods that personalize optimization-based controllers through learning-based components, enabling online parameter tuning. Examples include tuning MPC parameters for mobile robots with supervised learning~\cite{miyaoka2024chatmpc}, adjusting socially-aware navigation behavior via LLMs~\cite{martinez2025hey}, and modifying single-agent racing controllers~\cite{BaumannN2025enhacing}. Our work extends this direction in several important aspects, as we address motion planning in complex urban environments with dynamic multi-agent interactions, introduce a tunable risk-aware cost formulation explicitly designed for interaction-aware driving, and enable language-driven refinement of planner behavior.

Beyond personalization, explainability and natural interaction are emerging as critical requirements in autonomous driving systems. Recent efforts explore interpretable end-to-end driving via LLM-based reasoning \cite{xu2024drivegpt4} and conversational interfaces \cite{cui2024drive}. Our contribution advances this direction by incorporating a validation loop in which the LLM-generated behavioral modifications are translated into non-technical language and explicitly confirmed by the user prior to deployment, enhancing transparency and mitigating misalignment in motion planning behavior personalization.
    \section{Problem Formulation and Preliminaries}
We model the ego-vehicle as a discrete-time nonlinear stochastic dynamical system of the form
\begin{equation}
    \boldsymbol{x}_{t+1}=\mathcal{F}(\boldsymbol{x}_t, \boldsymbol{v}_t) \quad \boldsymbol{v}_t \sim \mathcal{N}(\boldsymbol{u}_t, \Sigma),
    \label{eq:dynamics}
\end{equation}
where $\boldsymbol{x}_t\in\mathbb{X}\subseteq\mathbb{R}^{N_x}$ denotes the system state at time step $t$, $\mathcal{F}$ is the possibly nonlinear state transition function, and $\boldsymbol{v}_t$ represents a stochastic control input drawn from a Gaussian distribution centered at the nominal control $\boldsymbol{u}_t\in\mathbb{U}\subseteq\mathbb{R}^{N_u}$ with covariance $\Sigma$. 

At each time step $t$, the vehicle operates within a specific driving scenario characterized by contextual features ${s}_t \in \mathcal{S}$. In addition, the user may provide behavioral preferences expressed as natural language queries (i.e. "Please be quick", "I am feeling dizzy"). We denote by $q_t \in \mathcal{Q}$ the set of user queries accumulated up to time $t$, which encode desired modifications of the vehicle’s driving style. These scenario features and user queries jointly define a parameterized cost functional $\mathcal{J}_{s_t,q_t}$, which shapes the motion planner’s behavior. 
We describe the problem hereafter:
\begin{problem}\label{prob:chance}
Given the dynamics defined in \eqref{eq:dynamics}, scenario features $s_t$, and user queries $q_t$, we seek to find (i) a cost functional $\mathcal{J}_{s_t,q_t}$ that meets $s_1$ and $q_t$, and (ii) the optimal sequence of control inputs that satisfy
\begin{subequations}\label{prob:synthesis}
\begin{align}
\underset{\boldsymbol{x}_t\in\mathbb{X}, \boldsymbol{u}_t\in\mathbb{U}}{\arg\min}\quad &\mathcal{J}_{s_t, q_t}(\boldsymbol{x}_t, \boldsymbol{u}_t)
  \label{eq:prob1}\\
\text{s.t. }\quad 
  & \boldsymbol{x}_{t+1}=\mathcal{F}(\boldsymbol{x}_t, \boldsymbol{v}_t)\\
  & \boldsymbol{v}_t \sim \mathcal{N}(\boldsymbol{u}_t, \Sigma). 
\end{align}
\end{subequations}
\end{problem}

\subsection{MPPI formulation} \label{sec:MPPI}
Model Predictive Path Integral (MPPI) control is a sampling-based approach for stochastic optimal control that leverages importance sampling to iteratively refine control inputs \cite{mppi_williams, williams_information-theoretic_2018}. At each planning step, MPPI generates $K$ candidate control sequences $\{V_k\}_{k=1}^K$ by sampling perturbations around the nominal sequence. Each sequence is propagated through the dynamics over a finite horizon of length $T$, yielding corresponding state trajectories $\{X_k\}_{k=1}^K$. For each trajectory, a cumulative cost is computed according to a predefined objective function $S_k=\mathcal{C}(X_k)$. The sampled trajectories are evaluated using an importance-weighting scheme that assigns higher weights to lower-cost rollouts. Specifically, the weight associated with the $k$-th trajectory is given by
\begin{equation}
    w_k = \frac{1}{\eta} \exp \left(-\frac{1}{\beta}(S_k -\rho)\right), \quad \sum_{k=1}^{K}w_k =1, \label{IS}
\end{equation}
where $\beta>0$ is a temperature parameter, % controlling the selectivity of the weighting, 
$\rho = \min_k S_k$ is introduced for numerical conditioning, and $\eta$ is the normalization constant.
% ensuring that the weights form a probability distribution. 
The updated optimal control sequence is obtained as the weighted average of the sampled sequences, $U^* = \sum_{k=1}^K w_kV_k$. In a receding-horizon fashion, only the first control action $\boldsymbol{u}_0^*$ is executed on the system. At the subsequent time step, the optimization is repeated, typically using a shifted version of the previously optimized sequence as a warm start.

\section{Approach}
Our motion planning framework, presented in Fig.~\ref{fig:overview}, leverages the reasoning capabilities of an LLM in order to interpret $s_t$ and $q_t$ and tune the cost function of an MPPI planner. The MPPI planner runs in a high-frequency computing the commands that minimize $\mathcal{J}_{s_t, q_t}$. The LLM module changes the MPPI cost function on demand, whenever the vehicle is in a new scenario or a new query is received. The user validates the changes before they are applied, and can modify the applied behavior by requesting new queries. It is therefore involved in two different loops to give feedback to the motion planner, always communicating with it using natural language. The MPPI cost formulation is explained in Sec.~\ref{sec:cost}, the interactions with the LLM in Sec.~\ref{sec:LLM}, and the adaptation of human feedback during execution in Sec.~\ref{sec:feedback}.

\subsection{Cost Formulation}
\label{sec:cost}
A key component of MPPI is the design of the cost function $S_k$ associated with each sampled control sequence. We design our cost function as:
\begin{equation}\label{eq:cost}
   S_k =  \omega_t\mathcal{C}_\text{tracking} + \omega_s\mathcal{C}_\text{speed}+\omega_{st}\mathcal{C}_\text{steering}+ \mathcal{C}_{\text{control}} +\mathcal{C}_{\text{risk}}
   \end{equation}
where $\omega_t, \omega_s, \omega_{st} \geq 0$ are tunable weights. The tracking cost, $\mathcal{C}_\text{tracking}$, penalizes the lateral deviation between the ego-vehicle and a reference path, encouraging lane-keeping or lane-changing behavior depending on the proposed reference path. The speed cost, $\mathcal{C}_\text{speed}$, penalizes deviations from a desired reference velocity profile. The steering cost, $\mathcal{C}_\text{steering}$, regularizes the steering angle to discourage excessive curvature, while $\mathcal{C}_\text{control}$ penalizes control effort, i.e., longitudinal acceleration and steering rate respectively weighted by $\omega_a\geq0$ and $\omega_{str}\geq0$, thereby encouraging smooth and comfortable maneuvers. The risk term, $\mathcal{C}_\text{risk}$, captures the probability of collision between the ego vehicle and surrounding agents along each sampled MPPI trajectory. At prediction timestep $t$, the collision region of the ego vehicle is denoted by $\mathcal{V}_{t}$ and that of agent
$a \in \mathcal{A}$ by $\mathcal{D}^a_{t}$, each represented as the union
of circular approximations. The collision region between the ego vehicle and agent 
$a$ is defined via the Minkowski sum, $\mathcal{R}^{\circ,a}_{t}
=
\mathcal{V}_{t} \oplus \mathcal{D}^a_{t}$ which represents all relative positions that would result in a collision.

Let $\boldsymbol{y}^a_{t} \in \mathbb{R}^2$ denote the predicted position of agent $a$ at timestep $t$, and let $\tilde{p}\!\left(\boldsymbol{y}^a_{t} \mid \cdot \right)$ be the corresponding predictive probability density obtained from the prediction model. The marginal collision probability for agent $a$ is computed by integrating this density over the collision region,
\begin{equation}
\hat{P}\!\left(\boldsymbol{y}^a_{t}\right)
=
\int_{\mathcal{R}^{\circ,a}_{t}}
\tilde{p}\!\left(\boldsymbol{y}^a_{t} \mid \cdot \right)
\, d\boldsymbol{y}^a_{t},
\end{equation}
Assuming conditional independence between agents at a given timestep, the joint collision probability over all agents $\mathcal{A}$ is computed as
\begin{equation}
\label{eq:joint_collision}
\hat{P}_{\mathrm{joint}}\!\left(\boldsymbol{Y}_{t}\right)
=
1 -
\prod_{a \in \mathcal{A}}
\left(
1 - \hat{P}\!\left(\boldsymbol{y}^a_{t}\right)
\right).
\end{equation}
where $\boldsymbol{Y}_t$ denotes the joint state of the surrounding agents. To evaluate \eqref{eq:joint_collision} efficiently, we employ a parallelized Monte Carlo approximation \cite{trevisan2025dynamic}, enabling scalable computation of the collision risk across all MPPI samples. The risk cost is defined as 
\begin{equation}
C_{\mathrm{risk}}
=
\omega_{\mathrm{soft}} \,
\hat{P}_{\mathrm{joint}}\!\left(\boldsymbol{Y}_{t}\right)
+
\omega_{\mathrm{hard}}
\mathds{1}_\sigma
\left(
\hat{P}_{\mathrm{joint}}\!\left(\boldsymbol{Y}_{t}\right)
\right),
\end{equation}
where $\omega_{\mathrm{soft}}, \omega_{\mathrm{hard}} \geq 0$
are tunable weights. The first term, $\omega_\mathrm{soft}$ imposes a penalty on on the collision probability, encouraging the planner to consistently prefer lower-risk trajectories. The second term is an indicator function returning $1$ when the collision probability is above a chosen risk threshold $\sigma$, effectively discouraging trajectories that violate an acceptable safety margin, with a very high $\omega_{\mathrm{hard}}$ that can not be tuned. The threshold is set to apply the soft penalty only to trajectories where the collision probability is lower than 0.05, where collision is not likely to occur.

\subsection{LLM module}\label{sec:LLM}
The LLM module runs on demand and is decoupled from the MPPI control loop. The primary objective of this module is to interpret the scenario features $s_t$ and user queries $q_t$ and to adapt the cost function parameters $\boldsymbol{\omega}=\{\omega_t,\omega_s,\omega_{st}, \omega_a,\omega_{str},\omega_{\text{soft}}\}$ accordingly. To this end, we designed a structured prompt, represented in Fig.~\ref{fig:prompt}, to guide the LLM toward reliable and interpretable parameter tuning. %It clearly states the task to be solved by the LLM and provides enough context, as well as uses some practices that improve the LLM accuracy.

\begin{figure}[ht]
\centering
\begin{tcolorbox}[
    colback=gray!5,
    colframe=gray!75!black,
    title=LLM module internal prompt,
    width=\linewidth,
    sharp corners,
    boxrule=0.5pt,
    fontupper=\footnotesize
]
\textbf{System Prompt:} \\
You are tuning the MPC cost function of an autonomous vehicle. You can tune weights and floating point parameters. The vehicle can drive between 0 and $v_{max}$ m/s. The cost function contains the following weighted terms (0–10 integer weights): \texttt{[Weights description]}

You will be given:
\begin{itemize}
    \item Current weights
    \item Current floating parameters
    \item A scenario description
    \item A user objective or query (None if there is no user objective)
\end{itemize}

Examples of behavior rules: \texttt{[Behavior rules]}

You must respond in 4 steps:

\#1. Intention. Rewrite the driving objective in one paragraph, identifying what is the intended behavior and how the cost terms should be used. Format: M:\textless text\textgreater

\#2. Weight parameters. Provide ALL weights (including unchanged ones). Values must be integers from 0-10. Start with: \#2. Weight parameters.

\#3. Floating point parameters. Provide ALL floating point parameters. Start with: \#3. Floating point parameters.

\#4. Describe the changes. Explain the changes in simple human terms. Do not mention technical details such as weights, parameters or cost. Format: D:\textless text\textgreater

\textbf{User Prompt:} \\The current scenario description is: \texttt{[Description]}. Consider this information when tuning the weights.

\textbf{User Prompt:} \\The current values of the weights are: \texttt{[Weights value]}

\textbf{User Prompt:} \\The user query is: \texttt{[Query]}
\end{tcolorbox}
\caption{Prompt used in our LLM module. The \texttt{[brackets]} symbolize placeholders that are substituted by the weights descriptions and behavior rules in Fig.~\ref{fig:weights-desc}, the current weights value, the scenario description and the user query.}
\label{fig:prompt}
\end{figure}

\begin{figure}[ht]
\centering
\begin{tcolorbox}[
    colback=gray!5,
    colframe=gray!75!black,
    title=Weights Description,
    width=\linewidth,
    sharp corners,
    boxrule=0.5pt,
    fontupper=\footnotesize
]
\begin{itemize}
    \item \textit{velocity\_tracking}: Penalizes deviation from reference velocity. Higher → stronger speed tracking.
    \item \textit{input\_penalty}: penalizes acceleration and deceleration magnitude. Higher → smoother but less reactive. Lower → faster acceleration and braking allowed.
    \item \textit{steering\_rate\_penalty}: penalizes rapid steering changes. Higher → smoother steering. Lower → faster steering allowed.
    \item \textit{steering\_angle\_penalty}: penalizes large steering angles. Higher → prefers straight driving. Lower → sharper turns allowed.
    \item \textit{lateral\_deviation\_penalty}: penalizes deviation from lane center. Higher → resists lane deviations. Lower → allows more free movement.
    \item \textit{collision}: penalizes predicted collision risk. Higher → more conservative behavior. Safety-critical term.
\end{itemize}
\end{tcolorbox}
\begin{tcolorbox}[
    colback=gray!5,
    colframe=gray!75!black,
    title=Behavior Rules,
    width=\linewidth,
    sharp corners,
    boxrule=0.5pt,
    fontupper=\footnotesize
]
\begin{itemize}
    \item If strong acceleration is needed → increase velocity\_tracking and decrease input penalty.
    \item If the vehicle should probably be slow and be able to brake → decrease velocity\_tracking.
    \item If fast steering is needed → decrease steering\_rate penalty and steering\_angle penalty.
    \item If there are vehicles in the adjacent lane → increase lateral\_deviation penalty.
\end{itemize}
\end{tcolorbox}
\caption{Descriptions of the cost terms and behavior rules included in the LLM prompt.}
\label{fig:weights-desc}
\end{figure}

To facilitate correct interpretation, we include intuitive descriptions of each cost term and its behavioral implications, as depicted in Fig.~\ref{fig:weights-desc}, and a set of behavioral guidelines as in-context learning~\cite{dong2024survey}. To promote coherent reasoning prior to numerical parameter selection, we adopt a \textit{Chain-of-Thought} prompting strategy \cite{wei2022chain,kojima2022large}. The LLM is first instructed to provide a concise motivation explaining the intended behavioral adjustments. Only after this reasoning step, we ask the model to output the updated weight values in a predefined format, enabling automatic parsing and validation. Furthermore, we use a general purpose LLM, allowing the system to handle diverse natural language queries without additional fine-tuning. To ensure numerical stability and exploit the model's pre-trained ordinal reasoning capabilities, we discretize the cost function weight space of the LLM into an integer scale. Specifically, we define an integer weight vector, $\boldsymbol{\lambda}=\{\lambda_t, \lambda_s, \lambda_{st}, \lambda_a, \lambda_{str},\lambda_{\text{soft}}\}$ where each $\lambda_i \in \{1,...,10\}$ corresponds to one cost weight $\omega \in \boldsymbol{\omega}$. The actual continuous cost weights are computed as $\omega=\hat{\omega}\frac{\lambda}{\bar{\boldsymbol{\lambda}}}$, where $\bar{\boldsymbol{\lambda}}$ denotes the mean of the integer weights, and $\hat{\omega}$ is an expert-defined nominal scaling constant. This normalization preserves the relative importance of the weights while maintaining the overall magnitude of each cost term within a range suitable for stable MPPI optimization. Moreover, discretizing the weight space mitigates known floating-point tokenization artifacts in LLM outputs \cite{wallace2019nlp} and help query interpretation by aligning with the way LLMs are generally trained~\cite{ouyang2022training}. The LLM is also instructed to tune other hyperparameters that are part of the cost function. In our case, we only encode $v_{\text{ref}}$ as the reference velocity being tracked, but there could be other if considering other cost terms.

In addition to numerical parameters, the LLM is instructed to generate a concise natural-language description of the proposed behavioral modifications. This description is later presented to the user during the validation stage, enhancing transparency and interpretability. The full interaction history, comprising system prompts, user queries, and previous model responses, is concatenated and provided as input to the LLM at each invocation. This conversational memory enables iterative refinement: users may refer to previous adjustments, request additional modifications, or override earlier decisions. In this way, the system supports continuous, context-aware behavioral adaptation. The human-in-the-loop execution is further explained in the following section.

\subsection{User Interaction}\label{sec:feedback}
Based on the user input. the LLM module outputs a set of parameters that directly influence the optimization-based motion planner. However, accurately translating human intent into numerical cost weights is inherently challenging. Misinterpretations may arise due to ambiguities in the prompt, incomplete user specifications, or limitations in the model’s reasoning. To mitigate these risks and enhance alignment between intended and executed behavior, we close the loop with the user in two different ways.
\subsubsection{Validation Before Deployment}
Before the updated parameters are applied to the MPPI controller, the system enters a \textit{validation loop}, illustrated in Fig.~\ref{fig:overview}. In this loop, the user is presented only with the natural language description of the changes requested in the last step of the LLM query (Fig.~\ref{fig:prompt}), without specific information about the parameters or cost function terms. This description uses non-technical language, and conceptually explains the changes expected in the behavior of the vehicle. The aim of this loop is to ensure that the LLM has understood the query properly and find a confirmation of the changes from the user without requiring knowledge of control design. In case the user identifies inconsistencies or is not satisfied with the proposed changes, they may provide corrective feedback. This feedback is appended to the existing conversation history, which already includes the system prompt, user query, and LLM response. The LLM is then requested to complete the task again, but with its previous response and feedback to refine its answer. The process iterates until the user explicitly confirms the behavioral modification. Only then are the updated parameters applied to the motion planner. This iterative validation loop prevents unintended cost configurations from being deployed and increases user trust in the adaptive planning framework.

\subsubsection{Online Refinement During Execution}
Recent studies have raised concerns regarding the faithfulness and reliability of LLM-generated explanations~\cite{turpin2023language}. While our work does not explicitly solve this limitation, we mitigate its practical impact by enabling continuous user oversight and refinement. To do so, during vehicle operation, the system remains receptive to additional user queries at any time. New requests are incorporated into the ongoing conversation context, which includes all previous queries and LLM responses. This persistent conversational memory allows users to reference earlier instructions (e.g., “keep the smoother steering but increase responsiveness”) and enables the LLM to reconcile or combine successive behavioral adjustments when they are compatible.
    \section{Experiments}
We validated our proposed method using scenes from NuPlan~\cite{caesar2021nuplan}. We used GPT-4o-mini~\cite{hurst2024gpt} as the general purpose LLM (temperature 0), and \cite{wu2024smart} for surrounding vehicles trajectories predictions. Expert nominal parameters are set to $\hat\omega_s=2.2$, $\hat\omega_a=0.5$, $\hat\omega_{str}=0.1$, $\hat\omega_{st}=2.0$, $\hat\omega_t=300$ and $\hat\omega_{soft}=100$, hand-tuned for smooth and safe navigation. The MPPI controller is implemented in PyTorch, achieving control frequencies of 100 ms.

\subsection{Query-based Adaptation}\label{sec:exp-1}
\begin{figure}
    \centering
    \begin{tabular}{@{}c@{\hspace{0.1cm}}c@{\hspace{0.1cm}}c@{}}
       \adjustbox{trim={6cm 13cm 11cm 11cm}, clip, width=0.48\linewidth}{\adjustbox{rotate=-5}{\includegraphics{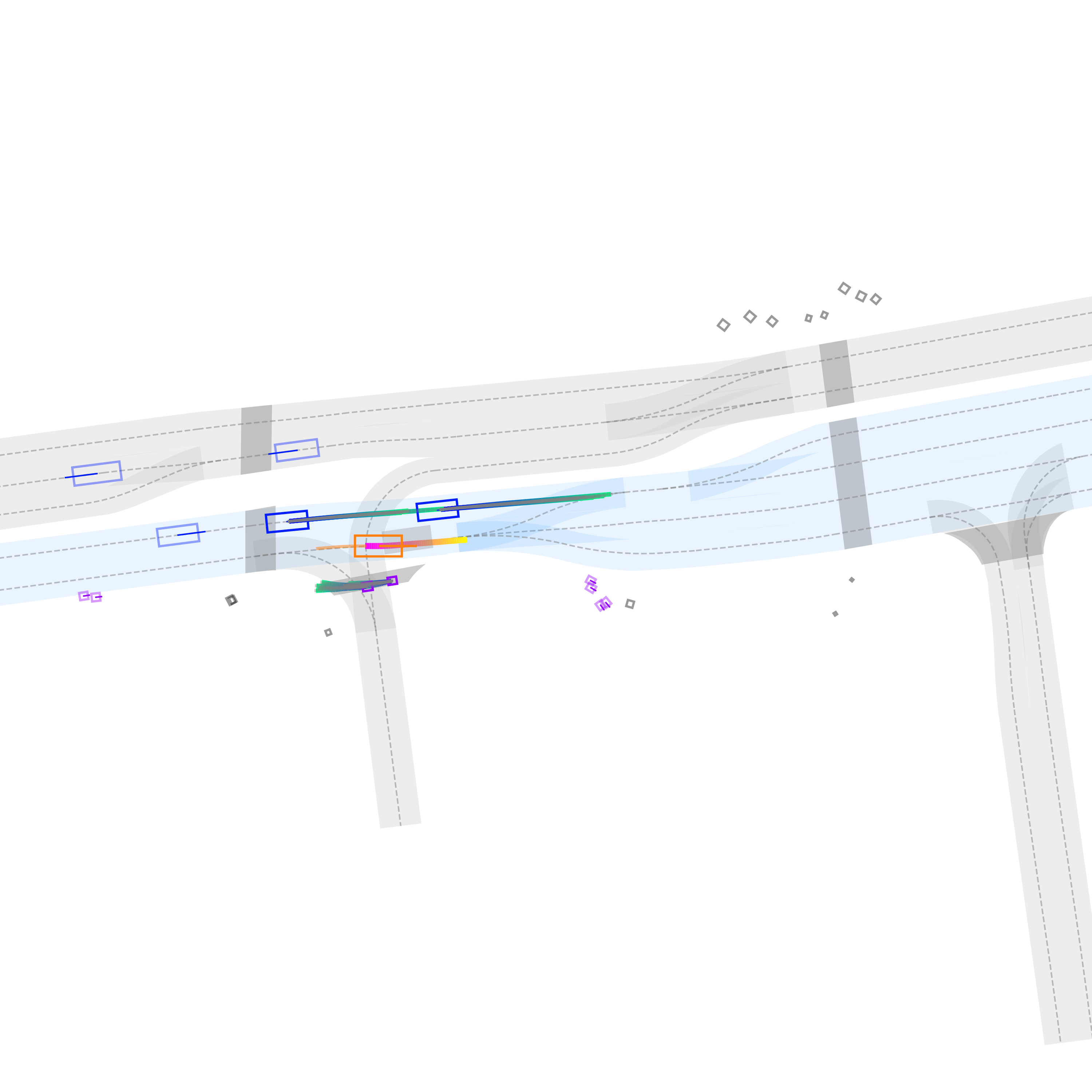}}}  & \adjustbox{trim={5cm 12.3cm 12cm 11.8cm}, clip, width=0.48\linewidth}{\adjustbox{rotate=5}{\includegraphics{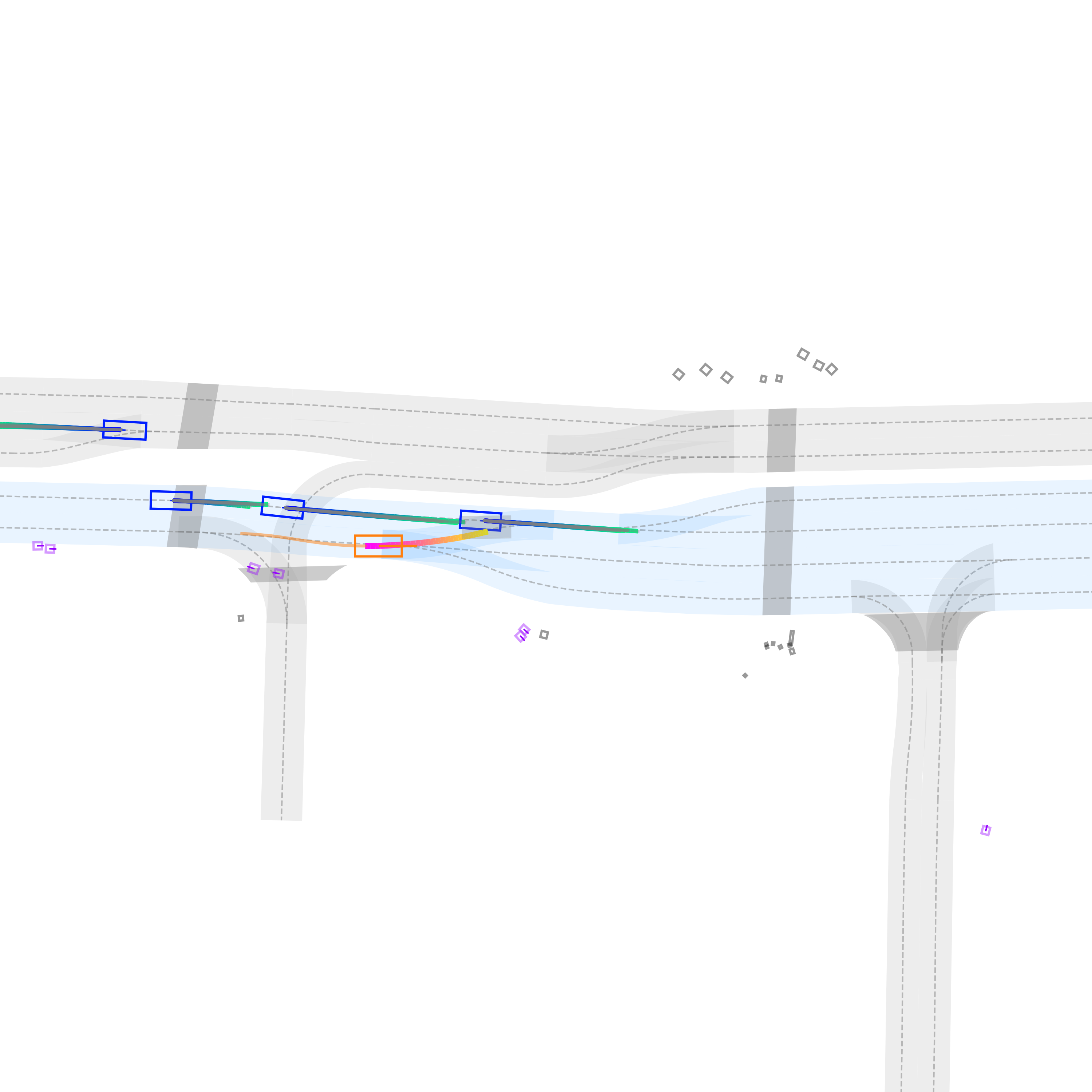}}} \\
        \footnotesize{(a) t: 3.0s} &
        \footnotesize{(b) t: 6.0s} \\
       \adjustbox{trim={7cm 13cm 14cm 14.6cm}, clip, width=0.48\linewidth}{\adjustbox{rotate=15}{\includegraphics{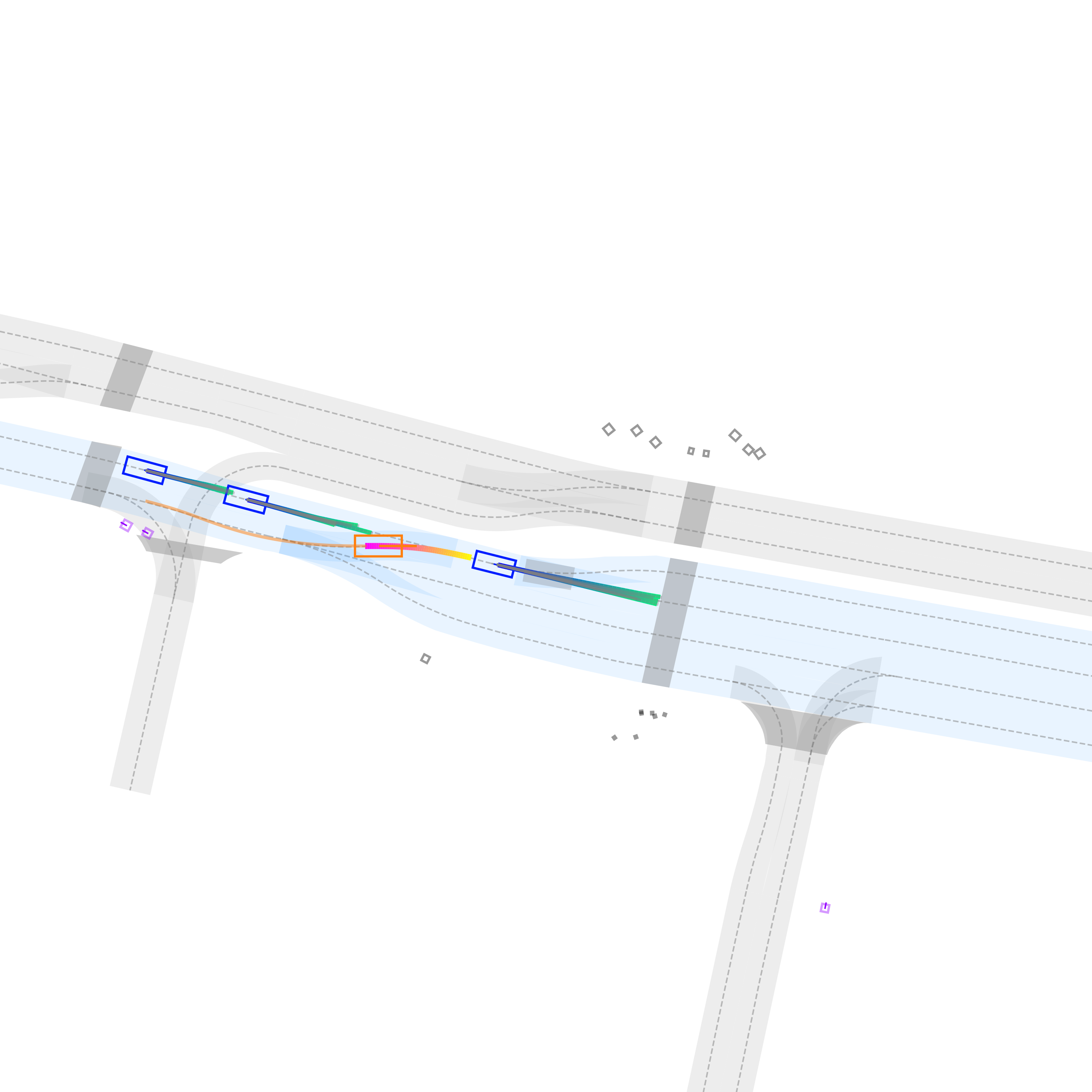}}}
       & \adjustbox{trim={3.8cm 11cm 11.8cm 11cm}, clip, width=0.48\linewidth}{\adjustbox{rotate=0}{\includegraphics{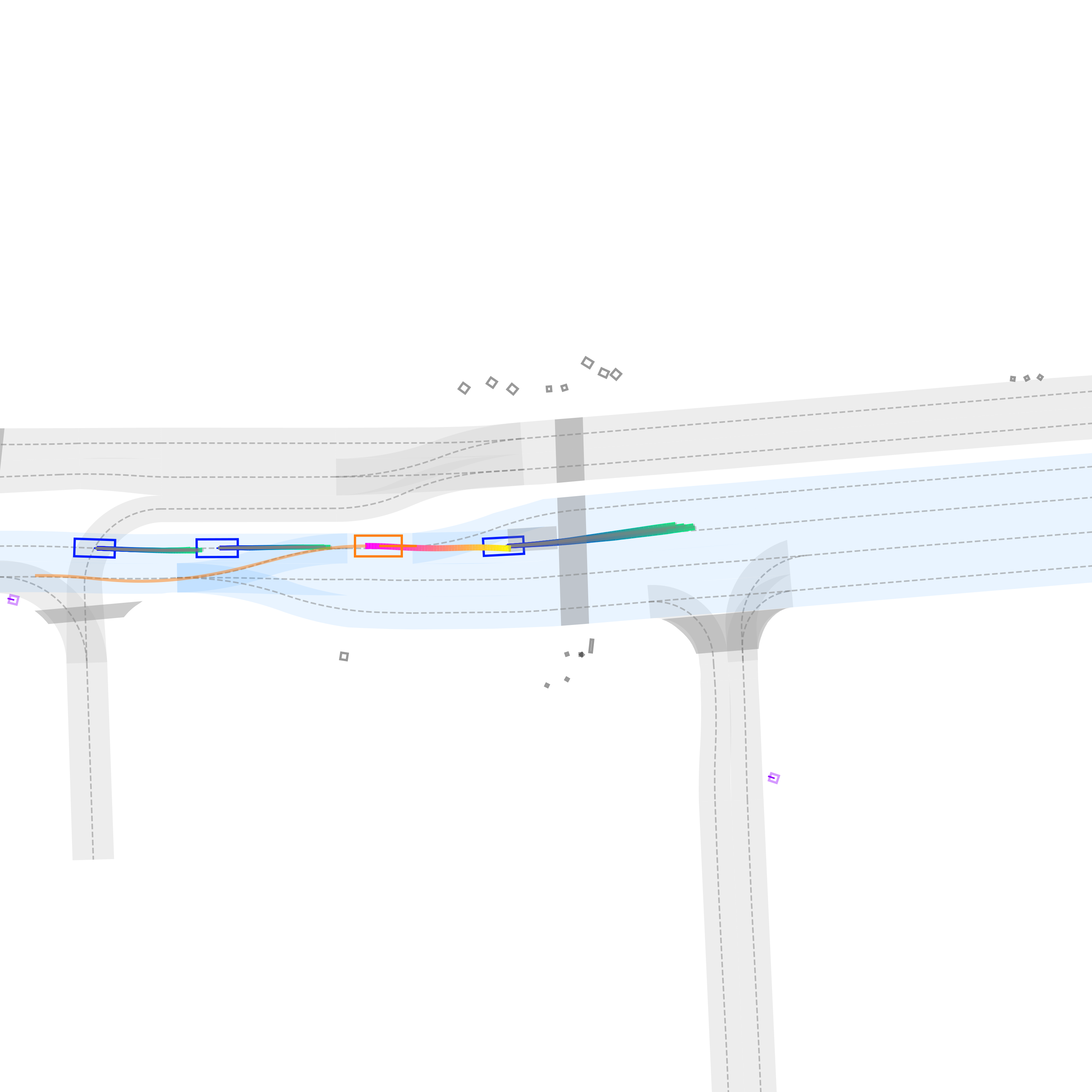}}}\\
         \footnotesize{(c) t: 9.0s} &
        \footnotesize{(d) t: 12.0s} 
    \end{tabular}
    \caption{The ego vehicle (orange) merges into an adjacent lane with surrounding traffic (blue). The orange line shows the past trajectory, the thick line the planned trajectory, and the lines extending from other vehicles their predicted motions.}
    \label{fig:merging}
\end{figure}

We used a cooperative merging scenario where the ego vehicle merges into an adjacent lane with interacting traffic (Fig.~\ref{fig:merging}) to evaluate the efficacy of our method in fulfilling different user-provided query specifications. We run the same scenario with different behaviors specified by user-provided queries, in Table~\ref{tab:queries-battery}, and gather metrics to quantitatively observe the behavioral changes of the vehicle. To evaluate conversational consistency, four experiments include sequential queries, preserving the full dialogue history when processing the second query. The corresponding query pairs are separated by dashed lines in Table~\ref{tab:queries-battery}. 

\begin{table}[ht]
    \centering
\begin{tabular}{c|c}
$\boldsymbol{q}$ & \textbf{Query} \\
\hline
$q_1$ & I'm feeling dizzy. \\
$q_2$ & I'm feeling better now, you can drive normally. \\
\hdashline
$q_3$ & I am running super late, be super fast. \\
$q_4$ & Keep it safe though, there is a lot of traffic. \\
\hdashline
$q_5$ & Be conservative. \\
$q_6$ & But don't be too slow. \\
\hdashline
$q_7$ & I am running late. \\
$q_8$ & We are very late, please be quick. \\
\end{tabular}
    \caption{Queries used for assessing the changes in the behavior of the vehicle, with dashed line separations corresponding to consecutive queries in the same experiment.}
    \label{tab:queries-battery}
\end{table}

\begin{table}[ht]
    \centering
\begin{tabular}{c|ccccccc}
$\mathbf{q}$ & $\mathbf{\lambda_s}$ & $\mathbf{\lambda_a}$ & $\mathbf{\lambda_{str}}$ & $\mathbf{\lambda_{st}}$ & $\mathbf{\lambda_t}$ & $\mathbf{\lambda_{soft}}$ & $v_{ref}$ \\
\hline
$q_1$ & 5 & 7 & 7 & 6 & 5 & 5 & 10 \\
$q_2$ & 7 & 5 & 5 & 5 & 5 & 5 & 10\\
\hdashline
$q_3$ & 7 & 3 & 4 & 4 & 5 & 5 & 15 \\
$q_4$ & 6 & 3 & 4 & 5 & 6 & 7 & 15 \\
\hdashline
$q_5$ & 4 & 6 & 6 & 5 & 6 & 7 & 10 \\
$q_6$ & 5 & 5 & 6 & 5 & 6 & 7 & 10 \\
\hdashline
$q_7$ & 7 & 3 & 4 & 4 & 5 & 6 & 15 \\
$q_8$ & 9 & 2 & 3 & 3 & 4 & 5 & 20 \\
\hline
$q_1-q_2$ & 6 & 4 & 5 & 5 & 5 & 5 & 10 \\
$q_3-q_4$ & 5 & 4 & 5 & 5 & 6 & 7 & 12 \\
$q_5-q_6$ & 5 & 5 & 5 & 5 & 6 & 6 & 10 \\
$q_7-q_8$ & 8 & 2 & 3 & 4 & 3 & 5 & 20 \\
\hline
$s_1$ & 3 & 5 & 5 & 5 & 6 & 8 & 10 \\
$s_2$ & 7 & 3 & 5 & 5 & 5 & 5 & 10 \\
\end{tabular}
\caption{Parameters resulting from the queries of the experiments in Sec.~\ref{sec:exp-1} (top), Sec.~\ref{sec:exp-2} (middle) and Sec.~\ref{sec:exp-3} (bottom).}
    \label{tab:queries-weights}
\end{table}

\begin{table}[ht]
    \centering
\resizebox{\linewidth}{!}{
\begin{tabular}{c|cccccc}
$\mathbf{q}$ & \textbf{St. rate} & \textbf{Acc.} & \textbf{St. angle} & \textbf{Speed} & \textbf{Tra. Dst.} & \textbf{Mrg. Time} \\
\hline
$q_1$ & 0.67 & 0.25 & 27.96 & 3.33 & 49.14 & 10.20 \\
$q_2$ & 0.62 & 0.22 & 28.68 & 3.50 & 51.67 & 10.00 \\
\hdashline
$q_3$ & 0.92 & 0.27 & 28.81 & 3.53 & 52.19 & 10.00 \\
$q_4$ & 0.62 & 0.24 & 28.54 & 3.48 & 51.41 & 10.10 \\
\hdashline
$q_5$ & 0.71 & 0.28 & 27.36 & 3.14 & 46.31 & 10.40 \\
$q_6$ & 0.61 & 0.22 & 27.67 & 3.26 & 48.15 & 10.30 \\
\hdashline
$q_7$ & 0.70 & 0.24 & 28.87 & 3.53 & 52.09 & 10.00 \\
$q_8$ & 0.85 & 0.28 & 29.05 & 3.52 & 52.08 & 10.00 \\
\hline
$q_0-q_1$ & 0.57 & 0.23 & 27.74 & 3.26 & 48.14 & 10.40 \\
$q_2-q_3$ & 0.69 & 0.21 & 28.09 & 3.39 & 50.00 & 10.10 \\
$q_4-q_5$ & 0.62 & 0.25 & 27.82 & 3.25 & 48.01 & 10.20 \\
$q_6-q_7$ & 0.90 & 0.27 & 29.05 & 3.54 & 52.25 & 10.00 \\
\end{tabular}
}
\caption{Metrics collected for the experimental queries, including (left to right): average absolute steering rate (rad/s), average absolute acceleration (m/s²), average absolute steering angle (deg), average speed (m/s), traveled distance (m), and merge completion time (s).}
    \label{tab:queries-metrics}
\end{table}

The parameters generated by the LLM module are reported in Table~\ref{tab:queries-weights}. The results demonstrate that the proposed framework systematically adapts the parameters in accordance with the specified queries. For instance $q_1$, which emphasizes smooth and comfortable driving, increases the relative importance of the acceleration, steering rate, and steering-related weights $(\lambda_a$, $\lambda_{str}$, $\lambda_{st})$. %, resulting in reduced control effort and smoother trajectories. 
When subsequently issuing $q_2$, the system rebalances the weights to recover nominal behavior, with increased emphasis on speed tracking. Similarly, query $q_3$ which requests more assertive behavior, reduces comfort-related penalties $(\lambda_a$, $\lambda_{str}$, $\lambda_{st})$ while increasing the speed weight $\lambda_s$, and the reference velocity $v_{\text{ref}}$. Query $q_4$ shifts the emphasis toward safety by increasing the soft risk weight $\lambda_{\text{soft}}$. Queries $q_5$ and $q_6$ prioritize responsiveness and comfort over speed, reflected in increased weights associated with risk and control smoothness, and slightly reduced reference velocity. In contrast, urgent queries $q_7$ and $q_8$ clearly prioritize efficiency and maneuverability, decreasing comfort-related penalties and increasing speed-related terms. 

The quantitative metrics in Table~\ref{tab:queries-metrics} corroborate these qualitative observations. Higher requested efficiency leads to increased average velocities, while comfort-oriented queries result in lower acceleration and steering rates. Moreover, the traveled distance during merging, used as an indicator of maneuver smoothness, decreases when conservative or comfort-oriented behavior is requested (e.g., $q_1$). It is important to note that the experiments were conducted in a realistic interactive setting where the ego vehicle must coordinate with surrounding traffic. As a result, the achieved average speeds are lower than the reference velocity $v_{\text{ref}}$, and merge completion times remain similar across experiments since the ego-vehicle must adapt its motion to surrounding traffic during the maneuvers, whose speed profile is kept identical across experiments.

\begin{figure}[!t]
\centering
\begin{tcolorbox}[
    colback=gray!5,
    colframe=gray!75!black,
    width=\linewidth,
    sharp corners,
    fontupper=\footnotesize
]
\textbf{User: } I'm feeling dizzy.\\
\textbf{LLM module:} The vehicle will drive more smoothly by making gentler turns and avoiding sudden accelerations or decelerations. This should help create a more comfortable ride and reduce any feelings of dizziness. Is this the behavior you intended with your query?\\
\textbf{User: } No, I'm feeling better now, you can drive normally.\\
\textbf{LLM module:} The vehicle will now drive in a more responsive manner, allowing for quicker acceleration and steering while still ensuring safety and comfort. This adjustment will help the vehicle adapt to normal driving conditions effectively.
\end{tcolorbox}
\begin{tcolorbox}[
    colback=gray!5,
    colframe=gray!75!black,
    width=\linewidth,
    sharp corners,
    fontupper=\footnotesize
]
\textbf{User: } I am running super late, be super fast.\\
\textbf{LLM module:} The vehicle will now focus on speeding up quickly and responding rapidly to driving conditions. It will allow for sharper turns and quicker lane changes, while still keeping an eye on safety to avoid any potential collisions. The target speed has also been increased to help achieve this faster driving objective. Is this the behavior you intended with your query?\\
\textbf{User: } No, keep it safe though, there is a lot of traffic.\\
\textbf{LLM module:} The vehicle will now drive faster than before but will still prioritize safety in heavy traffic. It will be more cautious with its speed and steering, ensuring it stays centered in the lane and avoids any potential collisions while still being responsive enough to navigate through traffic effectively. The target speed has been slightly increased to allow for a bit more urgency without compromising safety.
\end{tcolorbox}
\begin{tcolorbox}[
    colback=gray!5,
    colframe=gray!75!black,
    width=\linewidth,
    sharp corners,
    fontupper=\footnotesize
]
\textbf{User: } Be conservative.\\
\textbf{LLM module:} The vehicle will drive more cautiously, focusing on safety by reducing speed tracking and allowing for smoother acceleration and steering. It will also be more attentive to staying in the lane and avoiding any potential collisions, ensuring a safer driving experience overall. Is this the behavior you intended with your query?\\
\textbf{User: } No, don't be too slow.\\
\textbf{LLM module:} The vehicle will drive conservatively while still being responsive enough to react to the road conditions. It will focus on avoiding collisions and staying centered in the lane, but it will not sacrifice speed tracking or responsiveness, ensuring a balanced and safe driving experience.
\end{tcolorbox}
\begin{tcolorbox}[
    colback=gray!5,
    colframe=gray!75!black,
    width=\linewidth,
    sharp corners,
    fontupper=\footnotesize
]
\textbf{User: } We are very late, please be quick.\\
\textbf{LLM module:} The vehicle will focus on speeding up and being more responsive, allowing for quicker acceleration and a bit more freedom in steering. It will still be cautious about safety, but the priority is to drive faster to make up for lost time.
Is this the behavior you intended with your query?\\
\textbf{User: } No, We are very late, please be quick.\\
\textbf{LLM module:} The vehicle will prioritize speed and quick reactions, allowing for faster acceleration and more flexibility in steering. It will still keep an eye on safety, but the main goal is to drive as quickly as possible to make up for lost time.
\end{tcolorbox}
\caption{Illustration of the interactions of the user with LLM module in the validation loop of the experiments.}
\label{fig:exp-validation}
\end{figure}

\begin{figure*}
    \centering
    \begin{tabular}{@{}c@{\hspace{0.1cm}}c@{\hspace{0.1cm}}c@{\hspace{0.1cm}}c@{\hspace{0.1cm}}c@{}}
        \adjustbox{trim={4cm 9cm 12cm 11cm}, clip, width=0.195\linewidth}{\adjustbox{rotate=0}{\includegraphics{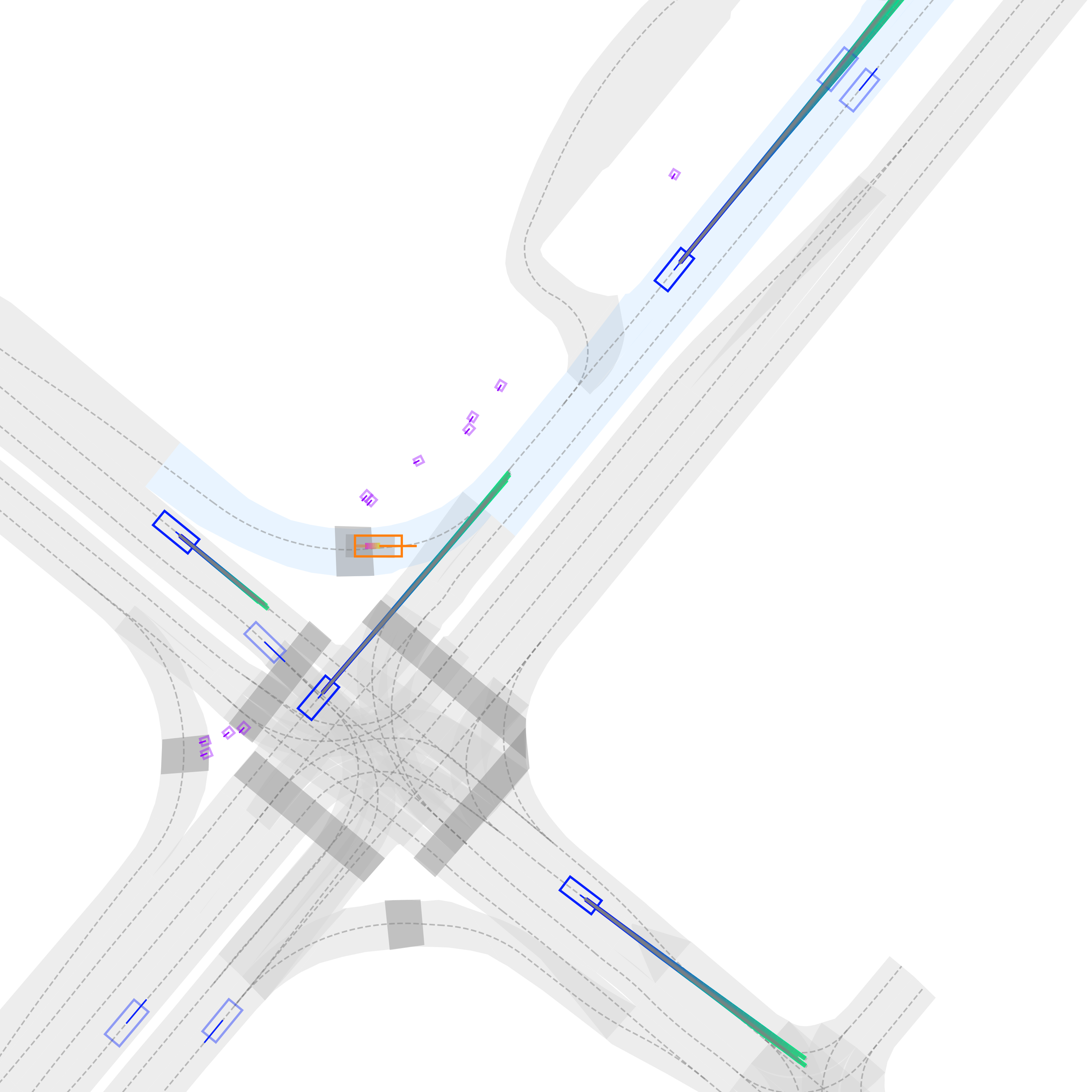}}} & 
        \adjustbox{trim={5cm 10cm 13cm 11cm}, clip, width=0.195\linewidth}{\adjustbox{rotate=0}{\includegraphics{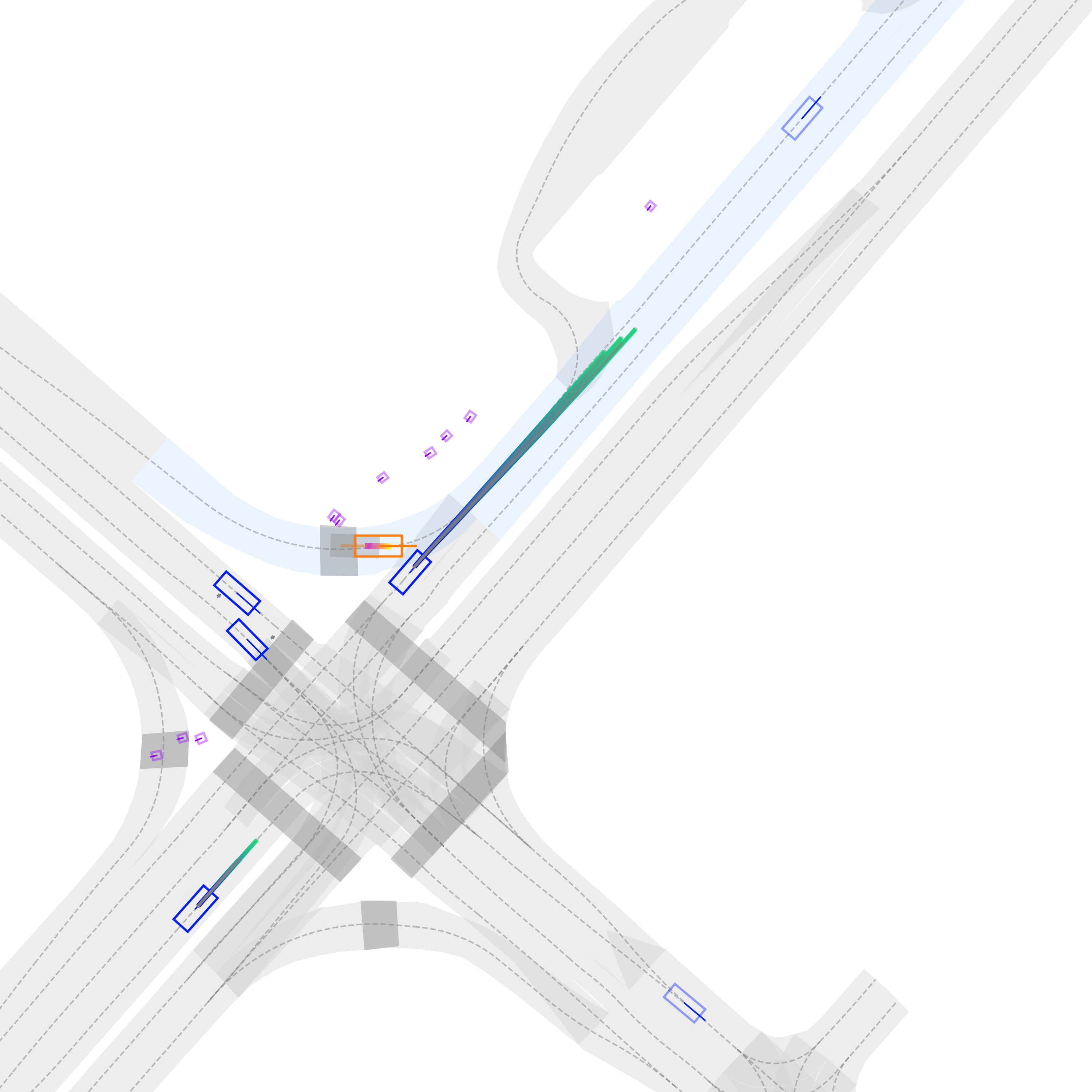}}} &
        \adjustbox{trim={5cm 10cm 13cm 11cm}, clip, width=0.195\linewidth}{\adjustbox{rotate=0}{\includegraphics{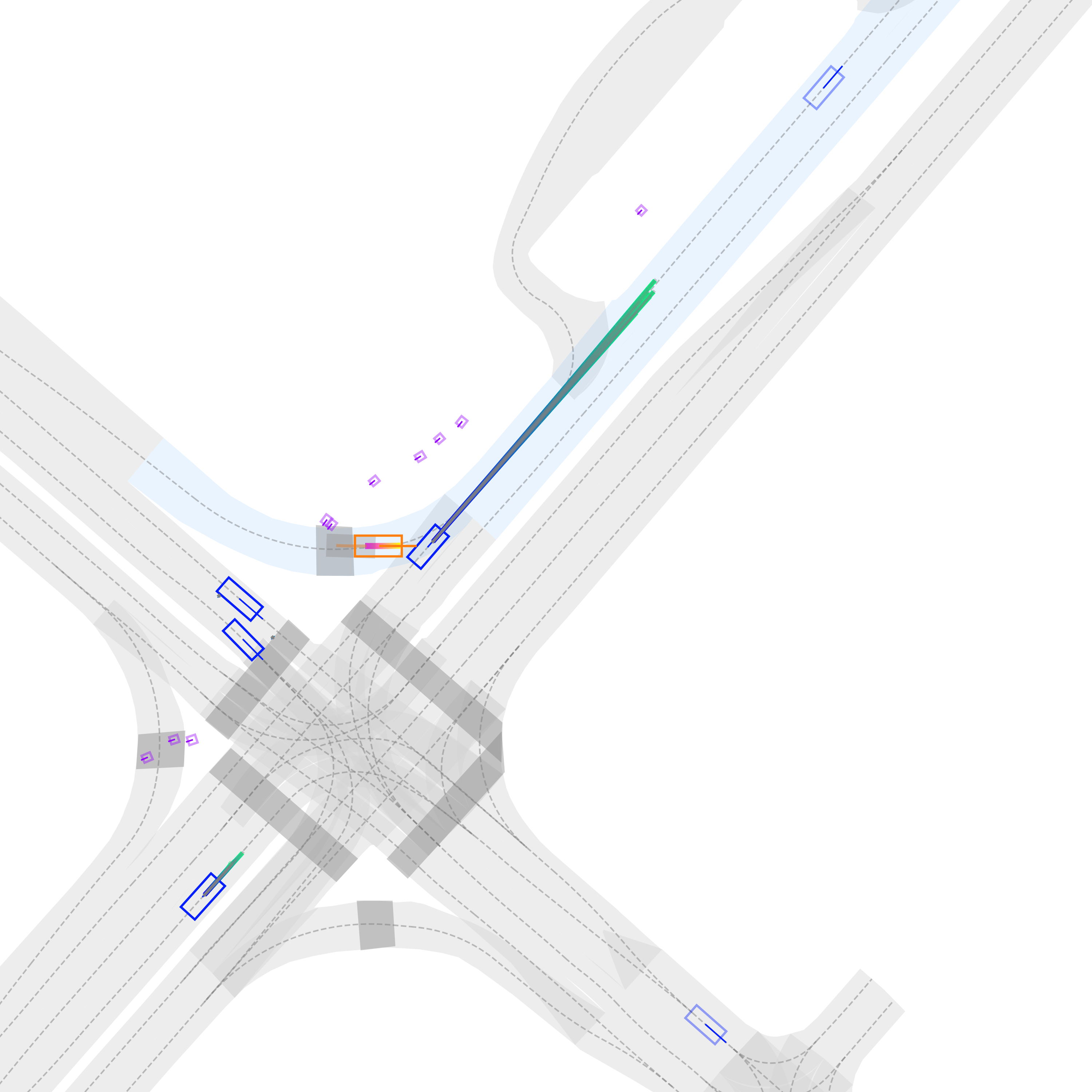}}} &
        \adjustbox{trim={4cm 11cm 12cm 10cm}, clip, width=0.195\linewidth}{\adjustbox{rotate=12}{\includegraphics{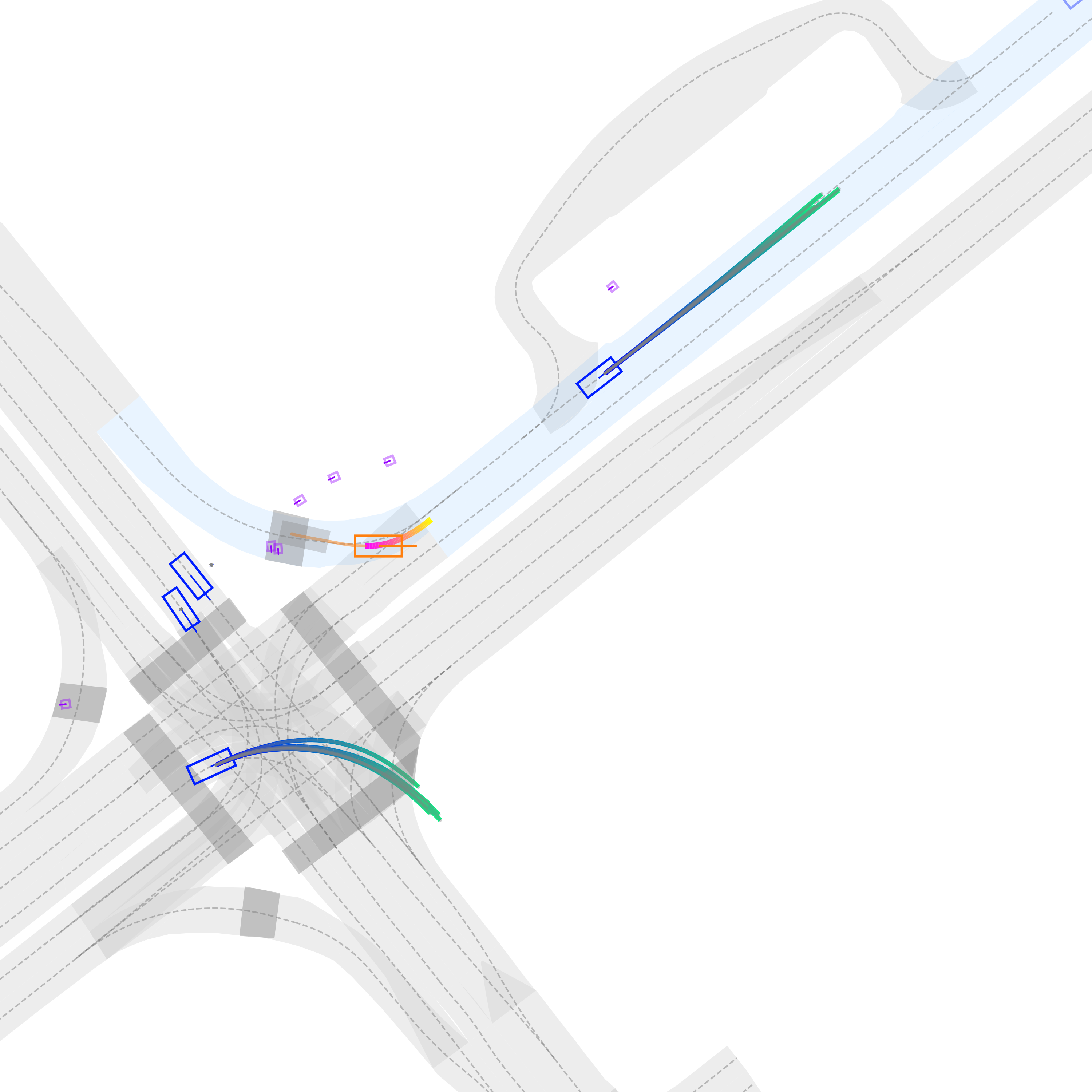}}} &
        \adjustbox{trim={8cm 12cm 16cm 16.5cm}, clip, width=0.195\linewidth}{\adjustbox{rotate=55}{\includegraphics{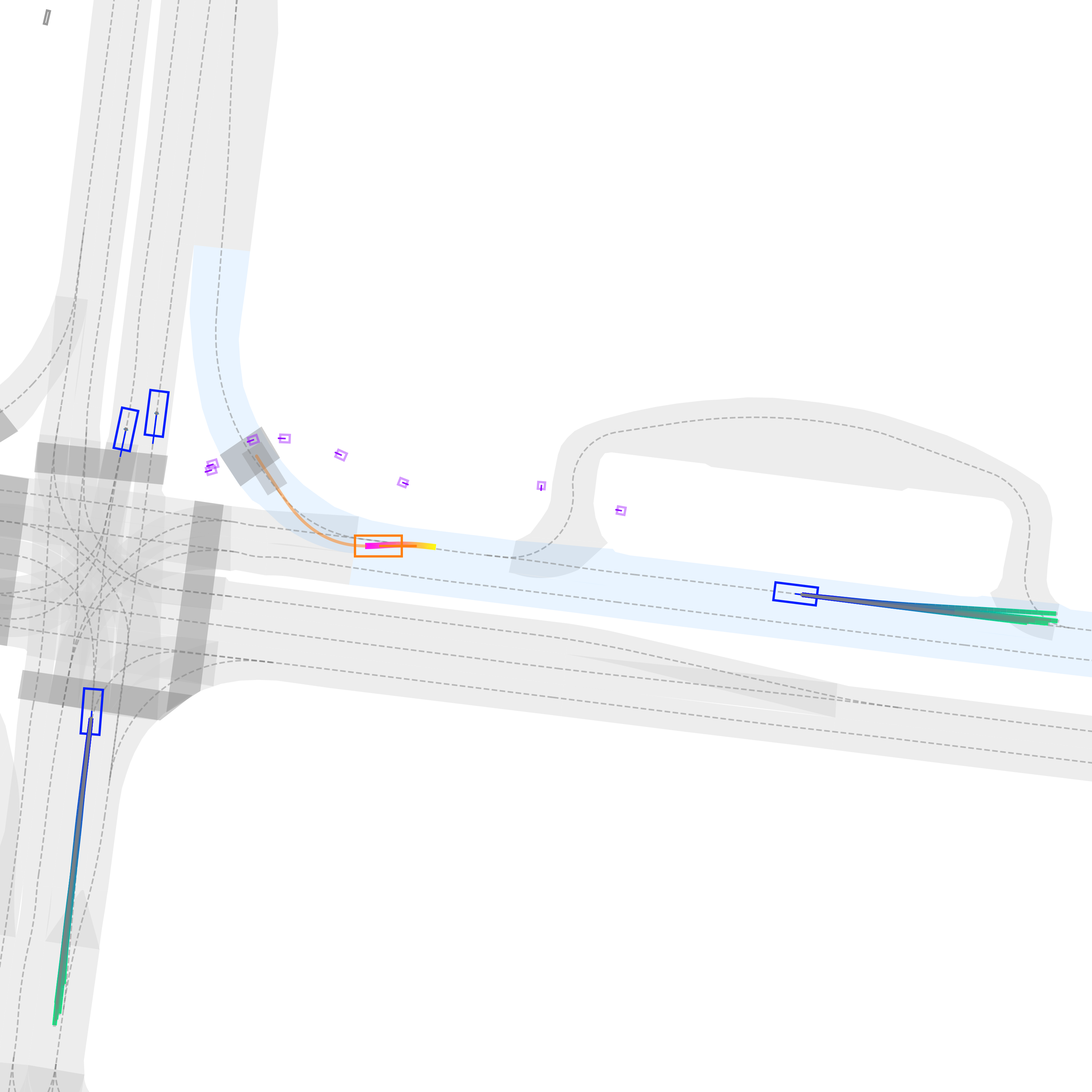}}}\\
        \footnotesize{(a1) t: 2.0s. $s_1$} &
        \footnotesize{(a2) t: 4.5s. $ s_1$} &
        \footnotesize{(a3) t: 5.0s. $s_1$} &
        \footnotesize{(a4) t: 9.0s. $s_1$} &
        \footnotesize{(a5) t: 12.0s. $s_1$} \\
        
        \adjustbox{trim={4cm 9cm 12cm 11cm}, clip, width=0.195\linewidth}{\adjustbox{rotate=0}{\includegraphics{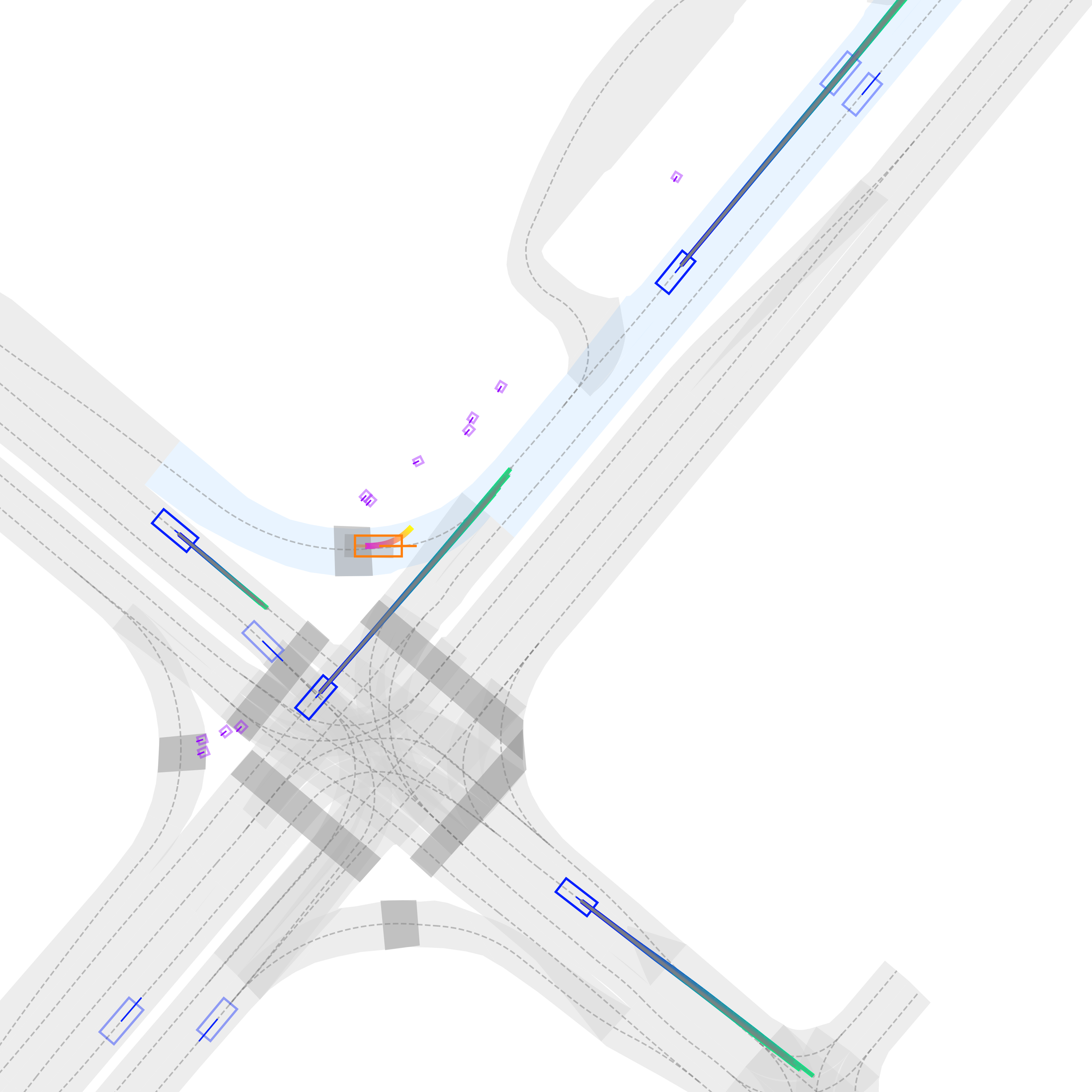}}} & 
        \adjustbox{trim={5cm 10cm 13cm 11cm}, clip, width=0.195\linewidth}{\adjustbox{rotate=0}{\includegraphics{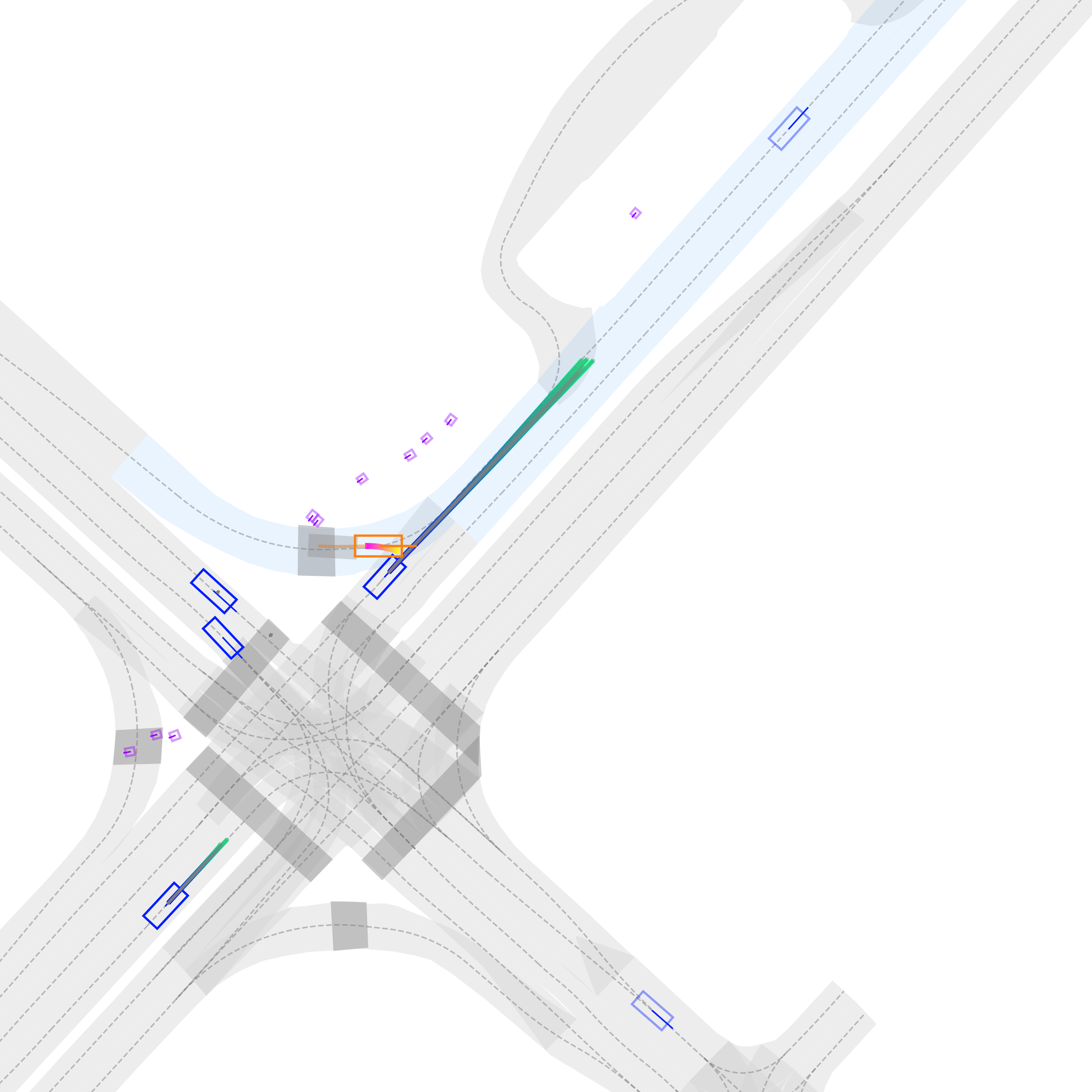}}} &
        \adjustbox{trim={5cm 10cm 13cm 11cm}, clip, width=0.195\linewidth}{\adjustbox{rotate=0}{\includegraphics{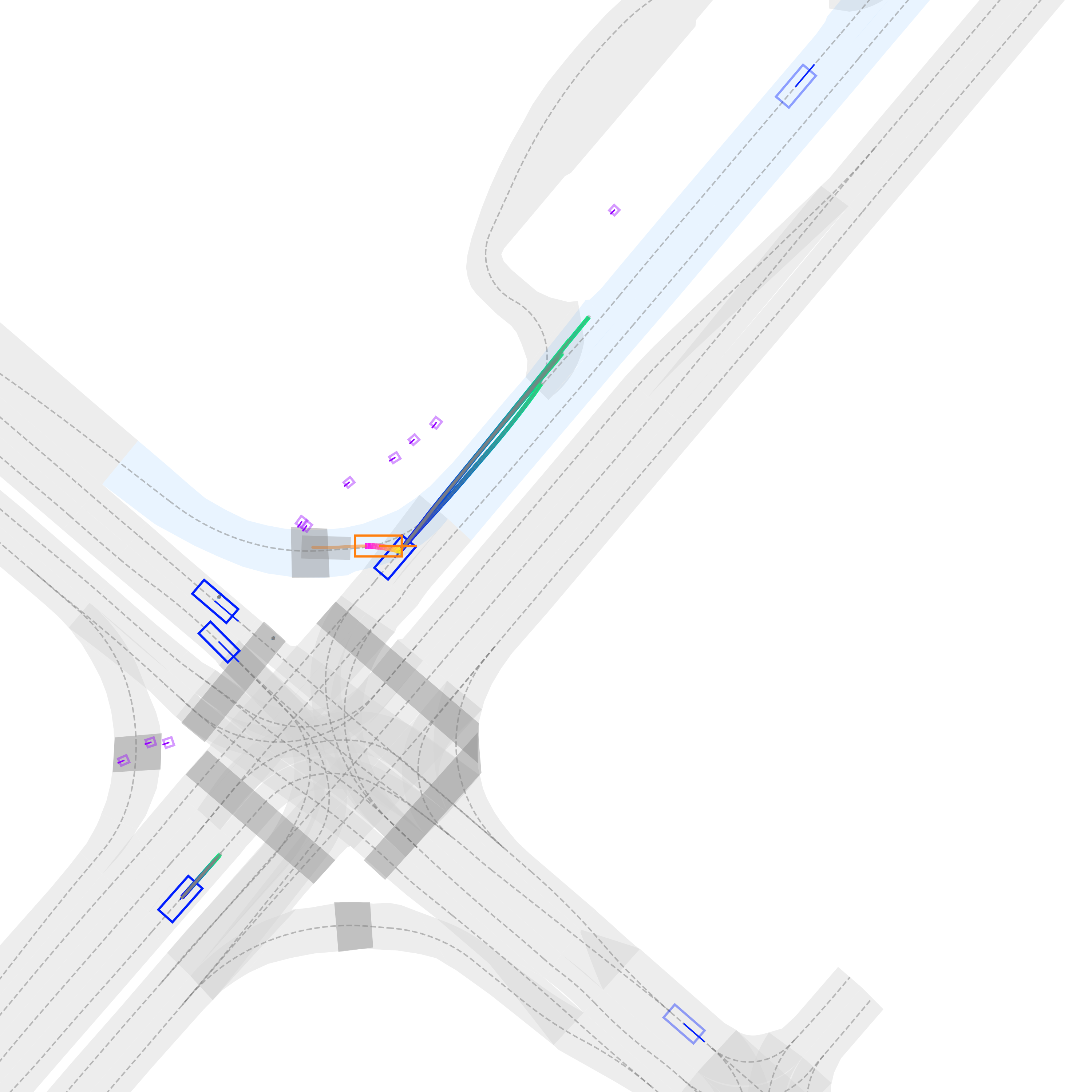}}} &
        \adjustbox{trim={4cm 11cm 12cm 10.5cm}, clip, width=0.195\linewidth}{\adjustbox{rotate=-10}{\includegraphics{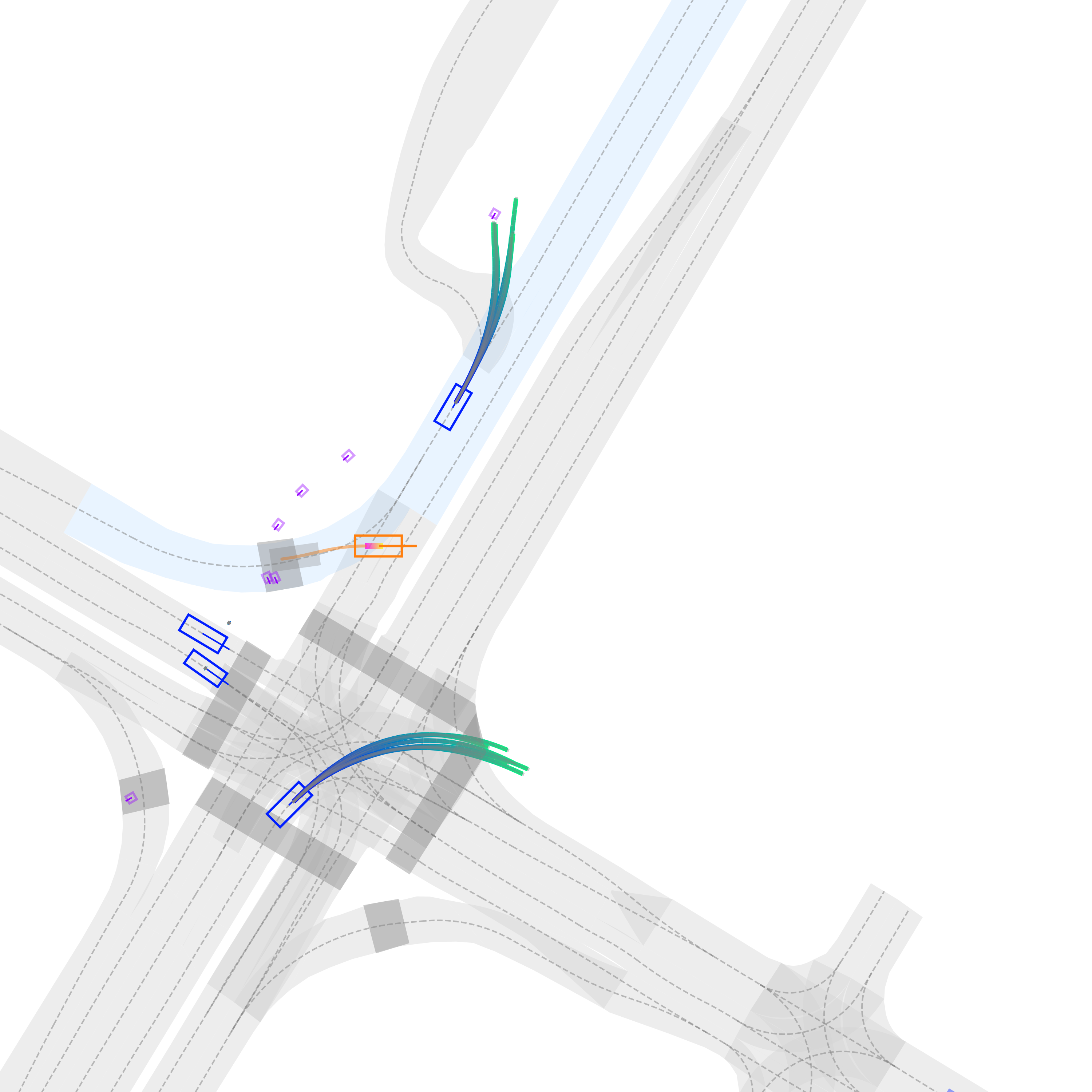}}} &
        \adjustbox{trim={5cm 12cm 13cm 10.5cm}, clip, width=0.195\linewidth}{\adjustbox{rotate=-10}{\includegraphics{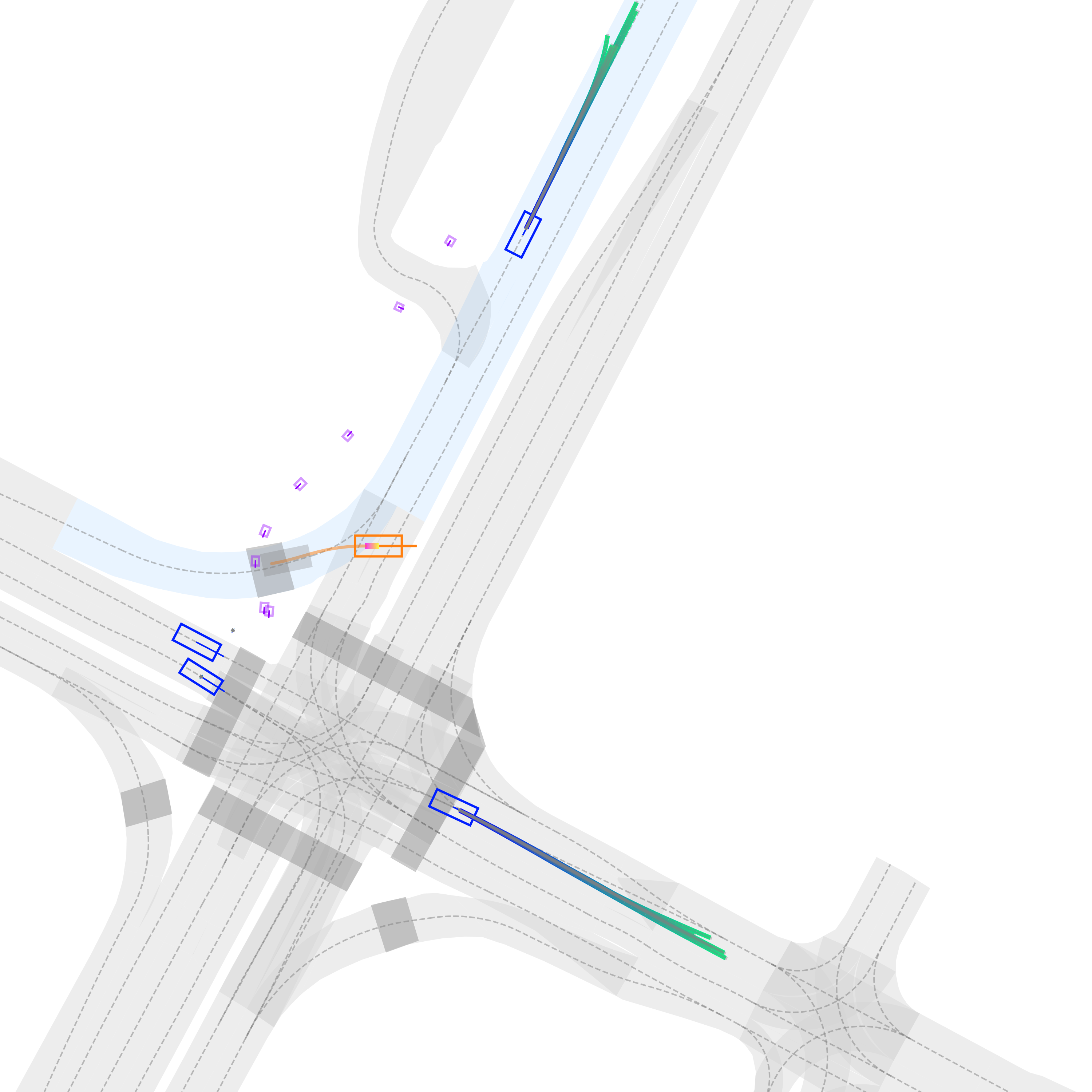}}}\\
        \footnotesize{(b1) t: 2.0s. $s_2$} &
        \footnotesize{(b2) t: 4.5s. $ s_2$} &
        \footnotesize{(b3) t: 5.0s. $s_2$} &
        \footnotesize{(b4) t: 9.0s. $s_2$} &
        \footnotesize{(b5) t: 12.0s. $s_2$} \\
        
        \adjustbox{trim={4cm 9cm 12cm 11cm}, clip, width=0.195\linewidth}{\adjustbox{rotate=0}{\includegraphics{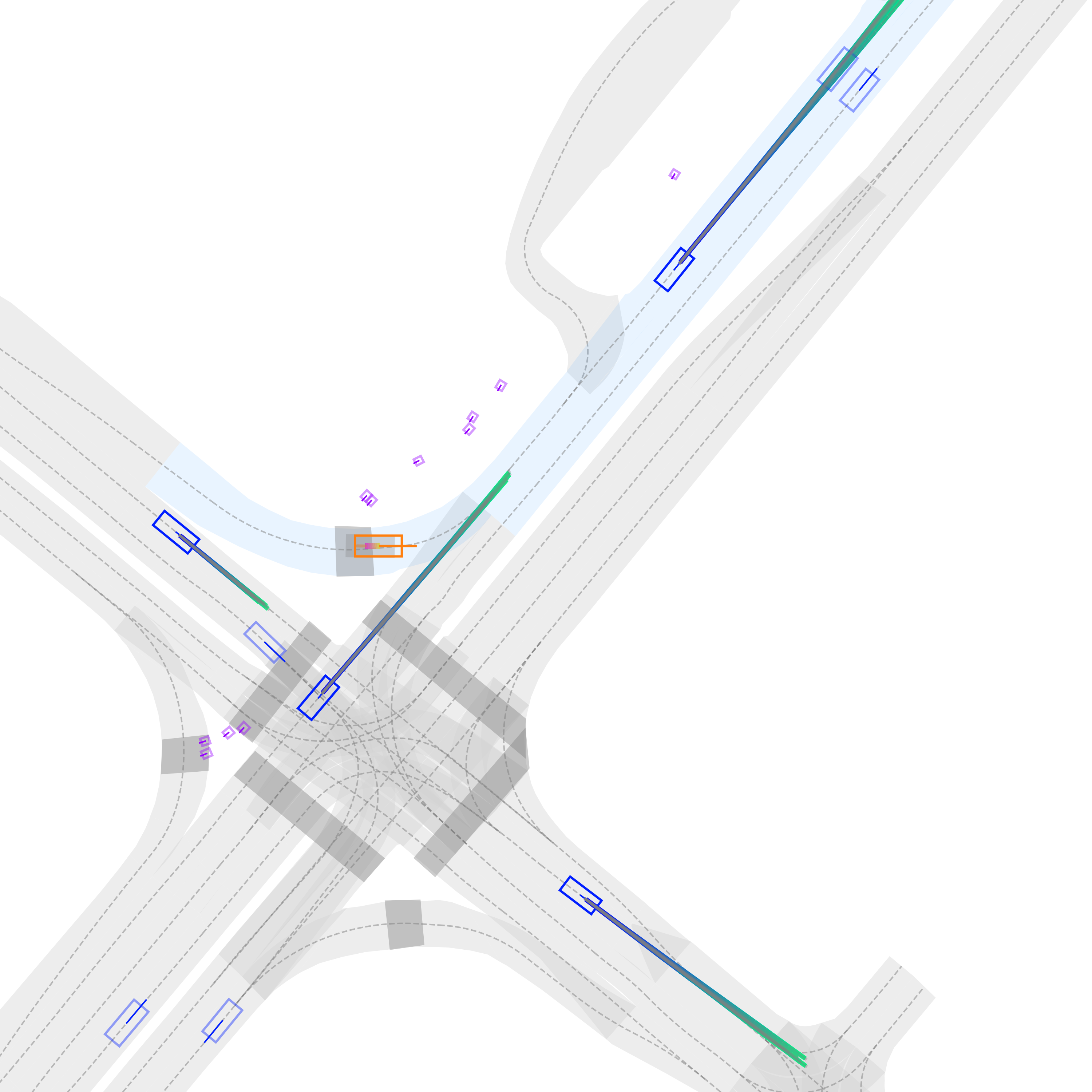}}} & 
        \adjustbox{trim={5cm 10cm 13cm 11cm}, clip, width=0.195\linewidth}{\adjustbox{rotate=0}{\includegraphics{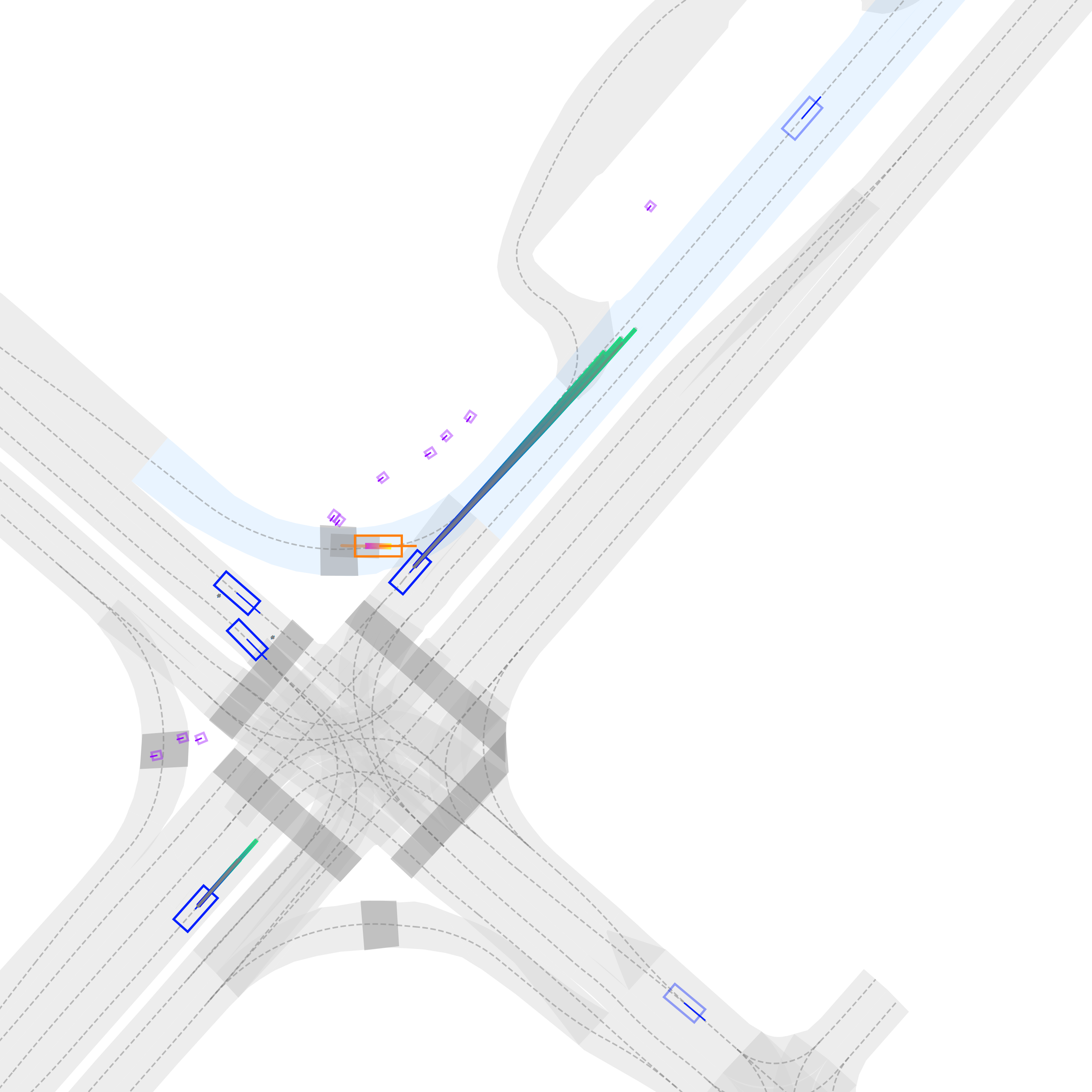}}} &
        \adjustbox{trim={5cm 10cm 13cm 11cm}, clip, width=0.195\linewidth}{\adjustbox{rotate=0}{\includegraphics{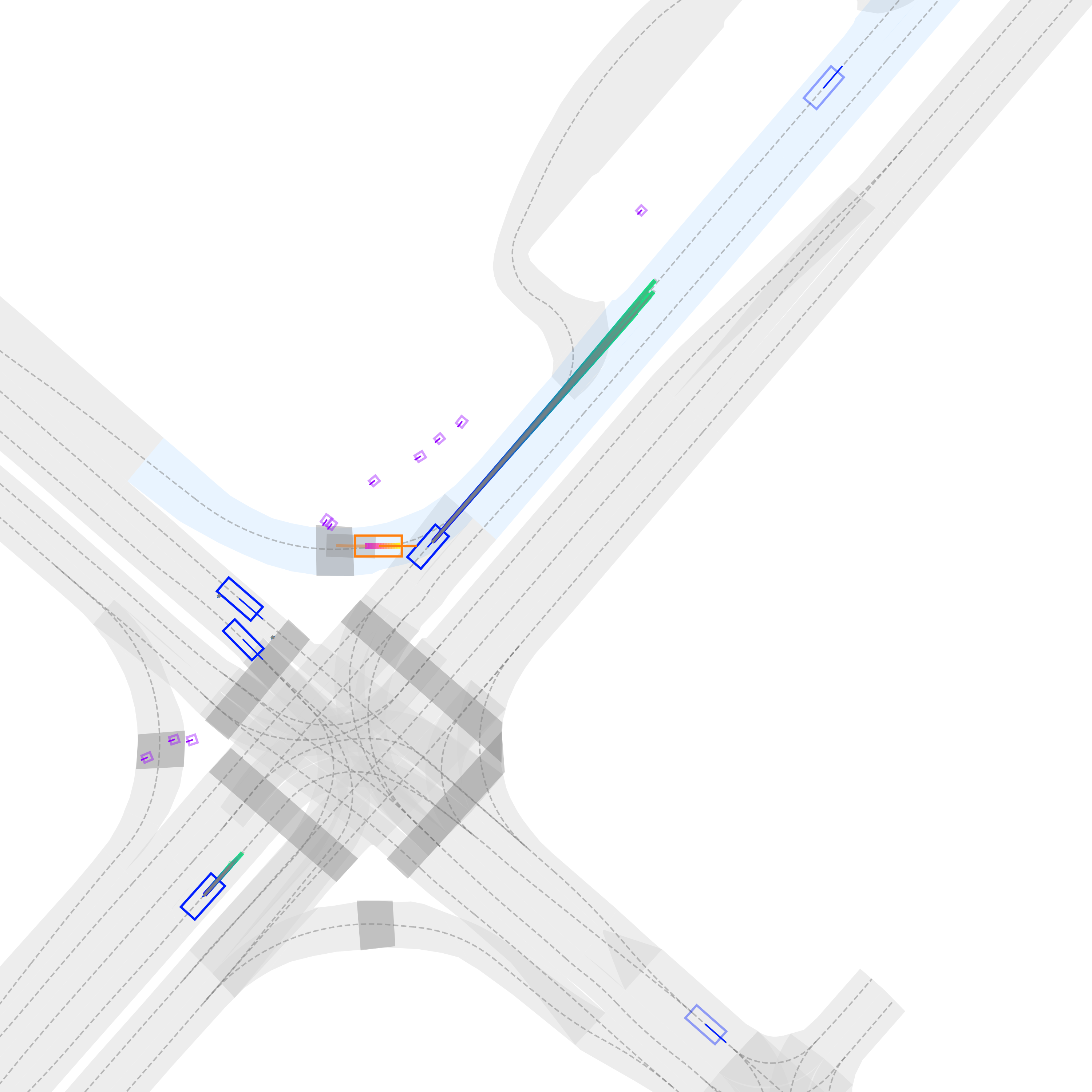}}} &
        \adjustbox{trim={4cm 11cm 12cm 9.5cm}, clip, width=0.195\linewidth}{\adjustbox{rotate=8}{\includegraphics{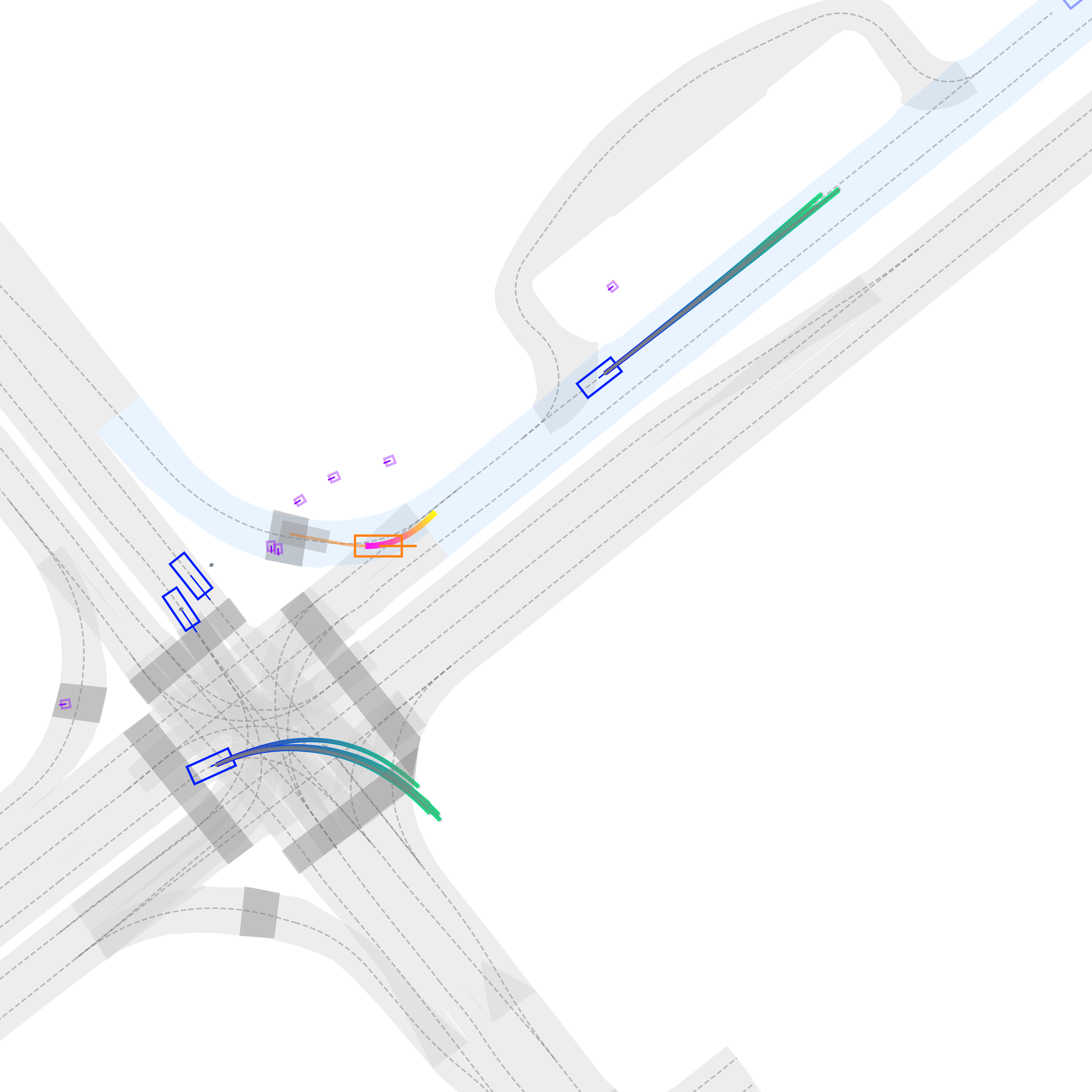}}} &
        \adjustbox{trim={8cm 12cm 16cm 16.5cm}, clip, width=0.195\linewidth}{\adjustbox{rotate=55}{\includegraphics{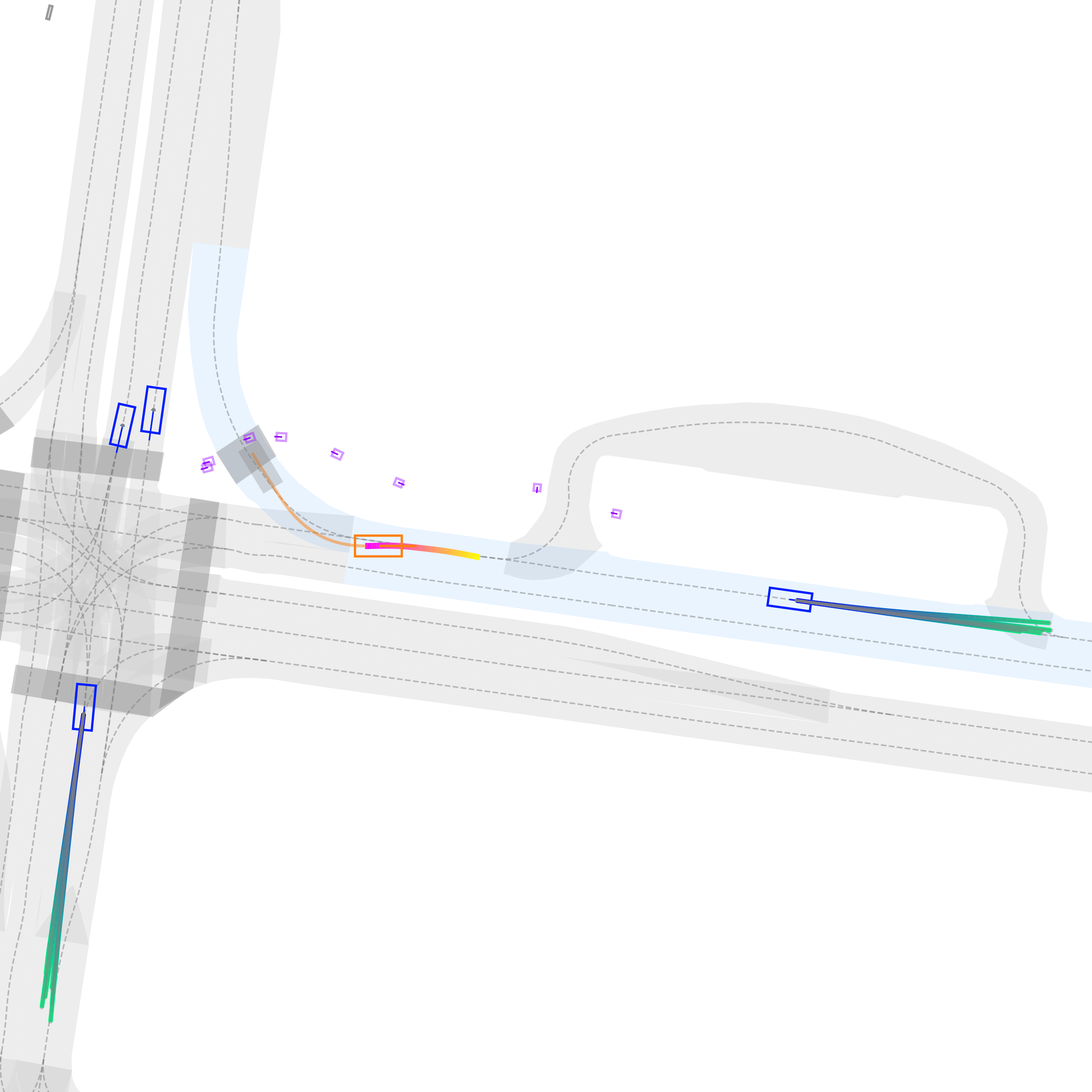}}}\\
        \footnotesize{(c1) t: 2.0s. $s_1-s_2$} &
        \footnotesize{(c2) t: 4.5s. $s_1-s_2$} &
        \footnotesize{(c3) t: 5.0s. $s_1-s_2$} &
        \footnotesize{(c4) t: 9.0s. $s_1-s_2$} &
        \footnotesize{(c5) t: 12.0s. $s_1-s_2$}  
    \end{tabular}
    \caption{Snapshots of a ramp-merge scenario where the vehicle faces a non-cooperative vehicle, using the 
    set of weights derived from the description $s_1$ (a), those derived from $s_2$ (b), and a combination of $s_1$ and $s_2$ depending on the scenario state (c). The plots follow the same representation as Fig.~\ref{fig:merging}.}
    \label{fig:exp-1}
\end{figure*}

\subsection{Validation Loop and Refinement}\label{sec:exp-2}
We next assess the effectiveness of the designed validation loop. The experimental setup is similar to that described in the previous section; however, in this case the second query is provided explicitly as feedback to correct or refine the LLM-generated proposal before it is applied to the motion planner. First, we show the user–LLM interactions for four representative experiments in Fig.~\ref{fig:exp-validation}. As shown, the LLM module first produces a concise and interpretable natural-language explanation of the proposed behavioral changes, summarizing its understanding of the user query and the resulting trade-offs. This explanation enables the user to assess whether the intended behavior aligns with their expectations. When the user provides corrective feedback, the LLM reasons about the implications of the new input in the context of its previous response and proposes a refined alternative. This iterative process can be repeated multiple times until the user decides that the proposal meets the expectations. Only then the corresponding parameters are applied to the motion planner, preventing unintended or misaligned behavior from being executed.

The parameters obtained after refinement and the corresponding performance metrics are reported in Tables \ref{tab:queries-weights} and \ref{tab:queries-metrics}, respectively. In these table, the notation $q_1-q_2$, for instance, denotes that the planner was first configured using query $q_1$ and subsequently refined using query $q_2$ through the validation loop. Interestingly, parameters and weights gathered after the refinement are similar to the ones obtained after applying the two queries consecutively, proving that both options could be used to update the vehicle behavior. However, the validation loop offers an additional safety and usability benefit by allowing users to correct misunderstandings \textit{before} execution, thereby reducing the risk of deploying unintended cost configurations.

\subsection{Scenario Specification}\label{sec:exp-3}
In the third set of experiments, we prove the efficacy of our proposed method to adapt to different scenario-level specifications. We set the vehicle in a different and challenging ramp-merge scenario, in which the ego vehicle must first regulate its speed to yield to a non-cooperative vehicle and subsequently merge into the main lane while accelerating to match traffic flow. This scenario highlights an inherent trade-off between conservative interaction at the beginning of the maneuver and assertive acceleration once a safe gap is available. 
We generated two distinct sets of weights using two different high-level scenario descriptions that are intentionally generic and not tied to a specific geometric configuration:
\begin{itemize}
    \item $s_1$: ``\textit{The vehicle must prioritize safety over progress. It should be prepared to stop at any time, avoid entering conflict zones without a clear and safe gap, and do not invade the space of other vehicles. Maintaining motion is not a priority.}"
    \item $s_2$: ``\textit{The vehicle has safely joined the lane and the path ahead is clear. The priority is to quickly accelerate to match traffic speed and avoid impeding flow. Comfort, smoothness, and strict lane centering are secondary. No immediate collision risk is present.}"
\end{itemize}
When applying the weights derived from $s_1$, the vehicle exhibits significantly more conservative behavior. Even though the phrase “clear and safe gap” is subjective, the does not explicitly evaluate or parameterize the size of an acceptable gap, and this behavior emerges from the tuning of the parameters. As shown in Fig. 6(a), it maintains safe distances from the interacting vehicle during the yielding phase (see (a2)–ab3)). However, this configuration does not efficiently recover the reference velocity after merging (see (a5), where the planned trajectory is short due to slow speed), leading to reduced traffic flow efficiency. Conversely, using the weights generated from $s_2$ produces aggressive acceleration behavior. While this configuration prioritizes rapid velocity recovery, it proves insufficiently cautious during the interaction phase and results in a collision in (b3), as illustrated in Fig.~6(b). This happens because the query states that there is no risk of collision, so the cost function does not prioritize $\mathcal{C}_{\text{risk}}$ even before the car has merged the lane. Finally, our proposed framework, in (c), enables a context-aware combination of both behavioral modes. By associating the appropriate scenario description with each phase of the maneuver, either manually or via a perception-based scene classifier, the LLM module adjusts the cost parameters dynamically, yielding conservative interaction when necessary and assertive acceleration when safe. This avoids the need to manually fine-tune a dedicated set of weights for every possible situation.

   \section{Conclusion}
This paper presented a motion planning framework that integrates a Large Language Model (LLM) with a risk-aware Model Predictive Path Integral (MPPI) motion planner to enable adaptive autonomous driving behavior. The MPPI planner maintains high-frequency control execution, while the LLM, based on the provided context or user intent on demand, proposes modifications to the cost function parameters of the MPPI controller. A human-in-the-loop validation mechanism enables users to confirm and refine behavioral changes prior to deployment, supporting transparent and reliable personalization. Simulation results in interactive driving scenarios demonstrate that the proposed method systematically adapts motion behavior to user intent and contextual requirements, including variations in smoothness, aggressiveness, efficiency, and safety prioritization. By enabling structured natural-language interaction with the motion planning layer, the framework helps bridge the gap between intelligent vehicle control systems and non-expert end users.

The performance of our approach could be improved with a tighter integration with perception and scene-understanding modules, enabling automatic extraction of contextual scenario descriptions without requiring explicit textual input. Moreover, the framework could be extended toward a more complete navigation stack, incorporating safety filters and higher-level discrete decision-making.
	
	%%%%%%%%%%%%%%%%%%%%%%%%%%%%%%%%%%%%%%%%%%%%%%%%%%%%%%%%%%%%%%%%%%
	%\addtolength{\textheight}{-12cm}
	%\vspace{10mm}
	\bibliographystyle{IEEEtran}
	% Your .bib file here
	\bibliography{IEEEabrv,root} 
	
\end{document}